%% file: acl_latex.tex
\pdfoutput=1

\documentclass[11pt]{article}

\usepackage[]{acl}

\usepackage{times}
\usepackage{latexsym}

\usepackage[T1]{fontenc}

\usepackage[utf8]{inputenc}

\usepackage{microtype}

\usepackage{inconsolata}
\usepackage{hyperref}       
\usepackage{url}            
\usepackage{booktabs}       
\usepackage{amsfonts}       
\usepackage{nicefrac}       
\usepackage{microtype}      
\usepackage{xcolor,colortbl}         
\usepackage{inconsolata}
\usepackage{amssymb}
\usepackage{amsthm}
\usepackage{amsmath}
\usepackage{multirow}
\usepackage{graphicx}
\usepackage{enumitem}
\usepackage{bm}
\usepackage{url}
\usepackage{hyperref}
\usepackage{todonotes}
\usepackage{subfiles}
\newtheorem{definition}{Definition}
\newtheorem{lemma}{Lemma}

\usepackage{caption}
\usepackage{subcaption}
\usepackage{centernot}
\usepackage{csquotes}
\usepackage[ruled,lined,linesnumbered]{algorithm2e}
\newcommand{\addsum}[1]{%
  \mathstrut\smash{\begin{array}[t]{@{}r@{}}#1\end{array}}%
}

\title{Favi-Score: A Measure for Favoritism in Automated Preference Ratings for Generative AI Evaluation}

\author{Pius von D\"{a}niken 
\and Jan Deriu 
\and Don Tuggener 
\and Mark Cieliebak \\
Centre for Artificial Intelligence\\
ZHAW School of Engineering \\
\texttt{\{vode,deri,tuge,ciel\}@zhaw.ch} \\
}

\begin{document}
\maketitle
\begin{abstract}
Generative AI systems have become ubiquitous for all kinds of modalities, which makes the issue of the evaluation of such models more pressing. One popular approach is preference ratings, where the generated outputs of different systems are shown to evaluators who choose their preferences. In recent years the field shifted towards the development of automated (trained) metrics to assess generated outputs, which can be used to create preference ratings automatically. In this work, we investigate the evaluation of the metrics themselves, which currently rely on measuring the correlation to human judgments or computing sign accuracy scores. 

These measures only assess how well the metric agrees with the human ratings. However, our research shows that this does not tell the whole story. Most metrics exhibit a disagreement with human system assessments which is often skewed in favor of particular text generation systems, exposing a degree of favoritism in automated metrics. This paper introduces a formal definition of favoritism in preference metrics, and derives the Favi-Score, which measures this phenomenon. In particular we show that favoritism is strongly related to errors in final system rankings. Thus, we propose that preference-based metrics ought to be evaluated on both sign accuracy scores and favoritism.


\end{abstract}

\section{Introduction}\label{sec:intro}
With the rise of Generative AI for text, images, code, music, and other modalities, the question of how to evaluate the quality of the outputs of these systems is becoming more and more important. \emph{Preference Ratings} have become a ubiquitous strategy for this task~\cite{belz-kow-2010-comparing,deriu-etal-2020-spot,stiennon2020booksummppo,freitag2021wmt21metrics,freitag-etal-2022-wmt22-results}. The general idea is that the outputs of two generative systems are shown to a (typically human) rater who decides which output they prefer, or whether both outputs are of equal quality. The ratings are then aggregated over a test set to compute in how many cases the outputs of one system were preferred over the other and whether the difference is significant. 

Since human-based evaluations are expensive, there has been a growing movement towards training automated metrics to assess generated content.~\cite{celikyilmaz2020evaluation, deriu2021survey,yeh2021comprehensive}. So instead of showing the outputs of two generative systems to human raters, one uses an automatic metric and hopes that the evaluation will resemble the ratings of the humans.~\footnote{For this paper, we treat human ratings as gold-standard, but we are aware of the issues with human ratings~\citep{amidei-etal-2018-rethinking, belz2021reprogen}}  
This raises the question of how these automated metrics themselves ought to be evaluated. Currently, the most popular method is measuring the correlation of the metrics' ratings to human judgments, or in the case of preference ratings, \emph{sign-accuracy} on both the sample and system level is computed (i.e., on how many preference ratings do human and metrics agree, or on how many system-pairs do human and metrics agree)~\cite{kocmi-etal-2021-ship}. For some tasks, such as machine translation, we have seen that according to these, metrics are steadily improving~\cite{freitag-etal-2022-wmt22-results}.  

While both measures are important in the assessment of automated metrics, they only count \emph{how many} mistakes the metric typically makes, ignoring how these mistakes are distributed. That is, whether the mistakes are systematically skewed in favor of one generative system under investigation or another. To underscore the significance of this information, let us consider the assessment of two generative systems, denoted as $\pi_a$ and $\pi_b$, using a metric that exhibits only a 10\% error rate, which are all skewed towards $\pi_a$. This means that in 10\% of cases, the metric's ratings will deviate from human judgments and in all cases the deviation is in favor $\pi_a$. Furthermore, assume that according to humans $\pi_a$ is slightly worse than $\pi_b$, then the favoritism of the metric will cause $\pi_a$ to be rated as being better than $\pi_b$ according to the metric. 

Thus, the research question that we tackle in this work is how we can measure favoritism, and investigate its impact on evaluation.




\paragraph{Contributions.} This work has three main contributions:
\begin{itemize}[noitemsep,topsep=0pt]
    \item We motivate and formally define the problem of favoritism in automated preference metrics for generative systems.
    \item We introduce the Favi-Score, an easy-to-compute score to quantify the favoritism exhibited by an automated preference metric in comparison to human judgments.
    \item We apply the Favi-Score to a variety of text generation tasks,~\footnote{Note that the formal definition of the Favi-Score could also be applied to other modalities. Thus, being agnostic to the task/domain at hand (e.g., music, images, or code generation).} showcasing the interplay between the Favi-Score and existing measurements of preference-metric quality.
\end{itemize}

Our main finding is that favoritism causes mistakes in the rankings of systems according to the metric: even a metric with a high sign accuracy can lead to wrong evaluation if it has a strong favoritism. Or stated differently, a metric with low sign accuracy and no favoritism can still yield the correct ranking of systems.~\footnote{We provide an implementation of the Favi-Score: \url{https://github.com/vodezhaw/faviscore}} 


\begin{figure}[t!]
\small
\centering
 \begin{subfigure}[c]{0.49\textwidth}
 \centering
     \includegraphics[width=1\textwidth]{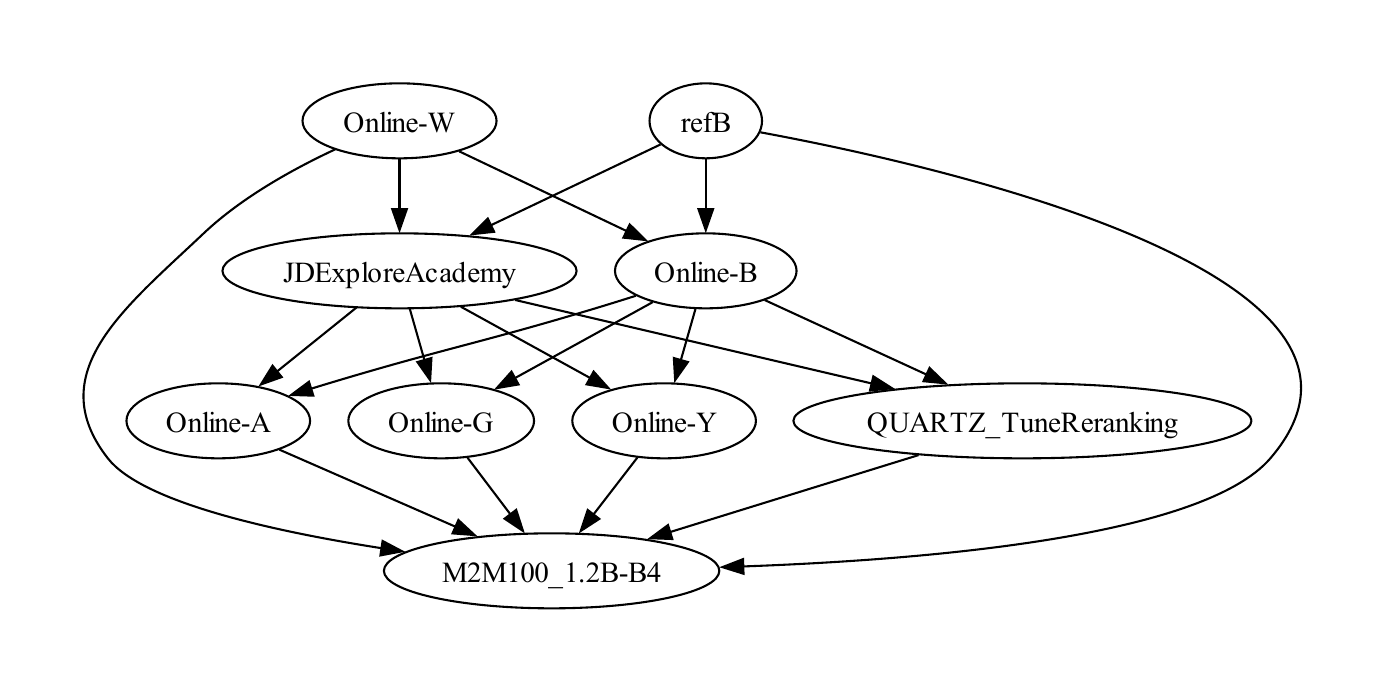}
     \caption{Human}
     \label{fig:daghum}
 \end{subfigure}
\begin{subfigure}[c]{0.49\textwidth}
 \centering
     \includegraphics[width=1\textwidth]{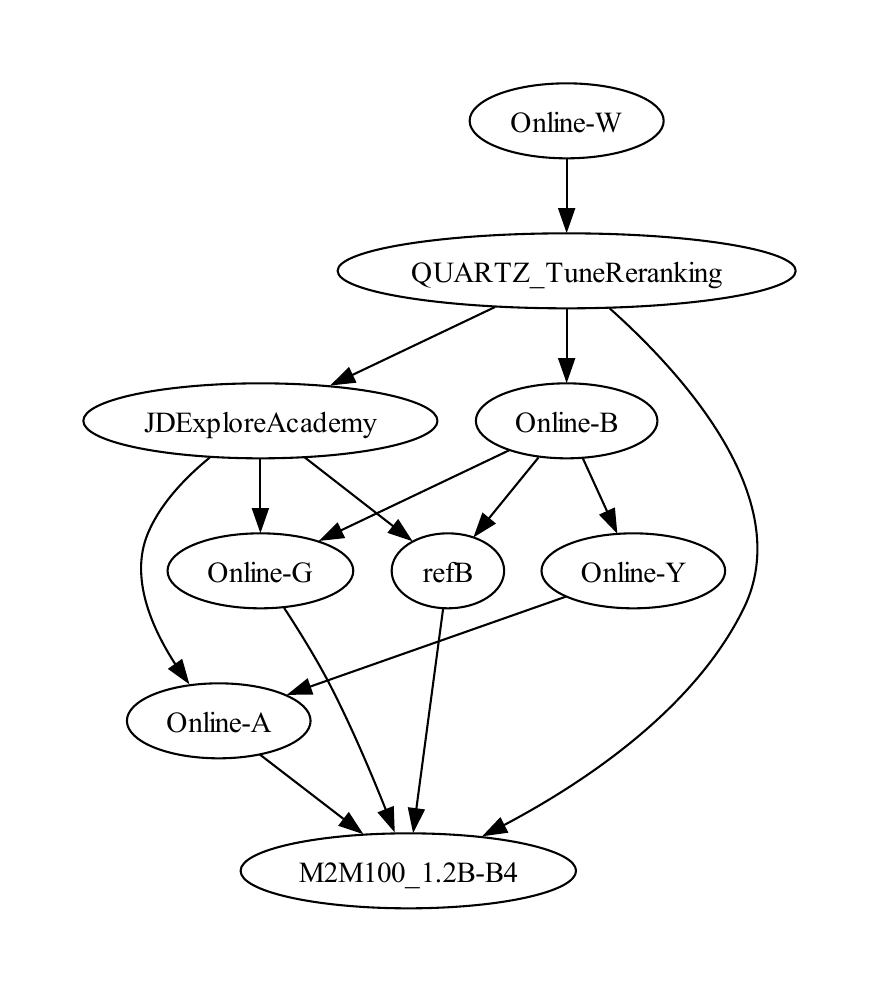}
     \caption{COMET-22-refA}
     \label{fig:dagcomet}
 \end{subfigure}
\caption{ Visualization of the consequences of unfair metrics by comparing the human ranking to the COMET ranking. The ranking of a set of systems is depicted as a Directed Acyclic Graph, where an edge from system A to system B states that system A "wins" against system B. Here, a win is determined by a sign test~\cite{sign_tests} at a 95\% confidence threshold.}
\label{fig:dag1}
\end{figure}

\section{Impact of Favoritism on System Rankings} \label{sec:mot_exp}
To demonstrate the need for the Favi-Score, we use the data of the WMT-22 metrics shared task~\cite{freitag-etal-2022-wmt22-results}. This contains ratings for 9 machine translation systems. Based on this data, we produce one ranking using human ratings and one ranking using COMET-22~\cite{rei-etal-2022-comet}, a well-established automated metric for machine translation outputs. We visualize their assessments using directed acyclic graphs (DAGs), as depicted in Figure~\ref{fig:dag1}. For this, the preference ratings are computed for each pair of systems using the human ratings and the corresponding COMET-22 ratings. Then, we put an edge between two systems if one system is statistically significantly preferred to the other system according to a sign test~\cite{sign_tests}. For better visibility, we omit transitive edges. Note that the system-level agreement lies at $61.1\%$, i.e., in $61.1\%$ the edges of the COMET-22 DAG correspond to the ones of the human DAG.

We see in Figure~\ref{fig:daghum} that according to humans, \emph{ref-B} and \emph{Online-W} are both at the top. However, in Figure~\ref{fig:dagcomet} according to COMET-22, \emph{ref-B} is ranked much worse, while \emph{QUARTZ\_TuneReranking} is disproportionately favored by COMET-22 compared to the human ranking. This is a direct consequence of favoritism. 

Note that \emph{only} considering the DAG does not reveal the whole story. For example, if a metric unduly favors a system that humans also rank highly, the overall ranking might not change. Yet, the favored system could appear to win by a much larger margin than it deserves, exaggerating its performance. For a concrete example, if we consider the outcome percentages between \emph{Online-W} and \emph{JDExploreAcademy}, then under human evaluation,  \emph{Online-W} is rated better in 24.1\% of cases and worse in 18.6\% of cases (giving a margin of 5.5\%). Under COMET-22, it is rated better in 57.5\% of cases and worse in 42.2\% (giving a margin of 15.3\%). Even though  \emph{Online-W} is rated better than \emph{JDExploreAcademy} in both cases, the perceived difference is much more pronounced under COMET-22, which is unfaithful to the human evaluation even though there is sign agreement. A concrete consequence is that, for example, a sign test will report much higher significance (lower p-value) for evaluation under COMET-22 than for human evaluation. This example shows the need for a mechanism that measures the inherent favoritism of a metric.  

Informally, the Favi score is a relative measure between the human ratings and the metric ratings. \emph{Instead of measuring how much the humans and metrics agree on their preference, it is a function of those samples where the humans and metrics disagree. The main idea is to count and weigh the cases in which the metric disproportionately assigns more samples in favor of one system than the human judges.} Thus, the Favi score yields a more fine-grained view compared to the DAG, as it also can measure the cases where there are no changes to the DAG, but one system is disproportionately favored (or disfavored) by the metric (see Examples in Appendix~\ref{app:examples}). The Favi score can be used as a diagnostic tool to discover those cases. In the subsequent sections, we will formally introduce the Favi score. 

\section{Related Work}
We are not aware of any prior work that defines or analyses favoritism in preference ratings. Most work on the analysis of metrics for text generation tasks is concerned with the overall performance, which is measured in terms of correlation to human judgments or sign accuracy~\cite{deriu2021survey,fabbri2021summeval,yeh2021comprehensive,freitag-etal-2022-wmt22-results}. The main conclusion of these analyses is that automated metrics do not yet match human ratings, and that the performance of the metrics vary depending on the domain and task. Moreover, there is a growing body of work that analyses individual metrics for generative tasks and their problems \cite[inter alia]{schluter-2017-limits,freitag-etal-2020-bleu,amrhein-sennrich-2022-identifying,hanna-bojar-2021-fine}, but these studies do not compare or assess different metrics.

\citet{novikova-etal-2017-need} go a step further and analyse specific weaknesses of various metrics for data-to-text. They showed that the metrics had difficulties distinguishing medium from good-quality outputs. Similarly, ~\citet{kryscinski-etal-2019-neural} showed that in the case of automated text summarization, automated metrics behave differently depending on whether the systems are extractive or abstractive. Thus, there is evidence in the literature that automated metrics are biased toward certain types of outputs and have difficulties with certain features. However, we are not aware of a fundamental analysis and discussion on the topic of structurally preferring the outputs of one system over others. 

A different line of work is concerned with adversarial attacks on metrics. \citet{sai2019re-eval-adem} showed that ADEM~\cite{lowe2017adem}, a trained metric, is susceptible to specific changes in the inputs.~\citet{deriu2022probing} devised a method based on reinforcement learning to find nonsensical adversarial samples that elicit perfect scores from metrics for conversational dialogue systems. These findings indicate a lack of robustness and raise the question why certain outputs are preferred. 

Finally, the last type of work related to ours is concerned with a more fundamental analysis of metrics. \citet{Wei2021TheSA} apply the bias-variance-noise decomposition from \citet{Domingos2000AUB} to assess statistical bias of automated metrics used in the evaluation of machine translation and summarization. Their analysis produces a ranking of automated metrics based on their statistical bias, but it does not indicate the direction of the bias. Also, their approach does not investigate the relation between the bias strength and the systems under scrutiny. While they find that automated metrics for Natural Language Generation (NLG) are biased, we extend this notion and show that the strengths of the biases vary on the system level and affects system rankings. \citet{chaganty2018debiasing} analyse the bias and variance of automated metrics. They find that automated metrics correlate poorly with human judgments, and that there is a correlation bias, i.e.\ the metrics often agree with humans in judging bad outputs but correlate poorly with humans on average or good quality outputs. 

To summarize, there is strong evidence that automated metrics exhibit behavior which can benefit or harm certain systems. To our knowledge, our work provides a definition and a measure for quantifying the amount of this benefit or harm for the first time.

\section{Preference Ratings}\label{sec:definitions}
In this section, we formalize the notion of a preference rating and define the confusion matrix which shows the amount of disagreement of two preference rating schemes (e.g.\ human judgments vs.\ an automated metric. 

We first start by formally defining a \emph{Generative System (GS)} as a function from an input space to an output space.
\begin{definition}[Generative System (GS)]
    Let $\mathcal{I}$ denote the set of all possible inputs to a GS, and $\mathcal{O}$ the set of all possible outputs. Then, a GS is a function $\pi$ that takes an input and generates an output 
     \begin{equation}
         \pi: \mathcal{I} \rightarrow \mathcal{O}
     \end{equation}
\end{definition}
Note that the input and output spaces are defined quite abstractly and are task-specific. For instance, in machine translation, the input space might be the set of utterances in the source language and the output space the set of utterances in the target language, whereas in news summarization, the input space would be the set of all news articles and the outputs is the set of all texts.

Next, we define the \emph{Preference Rating} as a function of a triple of one input and two outputs to a sign.
\begin{definition}[Preference Rating]
    Let $\mathcal{I}$ denote the set of all possible inputs to a GS, and $\mathcal{O}$ the set of all possible outputs. Then, we define a preference rating as:
    \begin{align}
        \mathcal{R}: \mathcal{I} \times \mathcal{O} \times \mathcal{O} \rightarrow \{+, =, -\}
\end{align}
where $+$ denotes that the first output is preferred over the second, $=$ denotes that both outputs are of equal quality, and $-$ denotes that the second output is preferred.
\end{definition}

Thus, for two generative systems $\pi_1$ and $ \pi_2$, $r[i] = \mathcal{R}(i, \pi_1(i), \pi_2(i))$ is the preference rating that compares the outputs of system $\pi_1$ and system $\pi_2$ for the input $i \in \mathcal{I}$. In the following, we denote a human-based preference rating as $r^H[i] = \mathcal{R}^H(i, \pi_1(i), \pi_2(i))$ and an automated metric-based preference rating as $r^A[i] = \mathcal{R}^A(i, \pi_1(i), \pi_2(i))$, respectively.   

In order to run an evaluation and apply a preference rating $\mathcal{R}$ to decide whether $\pi_1$ performs better than $\pi_2$, we first define the dataset that defines the evaluation setting. 

\begin{definition}[Evaluation Setting]
\label{def:eval_set}
    Given a test-set $\mathcal{T} \subseteq \mathcal{I}$, a pair of text generation systems $\pi_1$ and $\pi_2$, and two preference ratings $\mathcal{R}^H$ and  $\mathcal{R}^A$, we define the evaluation setting as:
    \begin{equation}
        \mathcal{E} = \{i, \pi_1(i), \pi_2(i), r^H[i], r^A[i] | i \in \mathcal{T} \}
    \end{equation}
\end{definition}

Thus, the evaluation setting $\mathcal{E}$ is composed of the inputs of the test set, the outputs of both TG systems, and the two preference ratings (e.g. according to humans and according to an automated metric). 

For a given evaluation setting $\mathcal{E}$, we can compute the confusion matrix between $\mathcal{R}^{A}$ and $\mathcal{R}^{H}$.

\begin{definition}[Confusion Matrix]\label{def:confusion_mat}
    Given an evaluation setting $\mathcal{E}$, we define the confusion matrix $\bm{C}$ as

\begin{equation}
    {\small
    \begin{aligned}
        \bm{C} = 
            \begin{pmatrix}
            C_{++} & C_{+=} & C_{+-} \\
            C_{=+} & C_{==} & C_{=-} \\
            C_{-+} & C_{-=} & C_{--} \\
        \end{pmatrix}
    \end{aligned}}
\end{equation}
where $C_{mn} = \sum_{i \in \mathcal{T}}\mathbb{I}[r^{H}[i] = m \land r^{A}[i] = n]$ and $\mathbb{I}$ is the indicator function.
\end{definition}

This corresponds to the usual confusion matrix known from classification settings, and the entries $C_{mn}$ count how many times a rating of type $m$ is classified as type $n$.

We call the cases where the automated rating disagrees with the human rating \emph{errors}.

\begin{definition}[Error]\label{def:error}
    Given an evaluation setting $\mathcal{E}$, $(i, \pi_{1}(i), \pi_{2}(i), r^{H}[i], r^{A}[i]) \in \mathcal{E}$ is an \emph{error} (of the automated rating with respect to the human rating) if $r^{A}[i] \ne r^{H}[i]$. 
\end{definition}

\begin{definition}[Total Error]\label{def:error}
    Given an evaluation setting $\mathcal{E}$, the the total error is given by $E = \sum_{i \in \mathcal{T}}\mathbb{I}[r^{A}[i] \ne r^{H}[i]]$. 
\end{definition}

To decide which of the two systems is better, we define the \emph{Outcome} of the evaluation, by counting the preferences. 

\begin{definition}[Outcome]
    Given an evaluation setting $\mathcal{E}$, we define the outcome $\bm{d}$ as
    \begin{equation}
    {\small
    \begin{aligned}
        \bm{d} = (d_{+}, d_{=}, d_{-} )
    \end{aligned}}
\end{equation}
where $d_n = \sum_{i \in \mathcal{T}}\mathbb{I}[r^{H}[i] = n]$. We call $mar(d) = d_{+} - d{-}$ the outcome margin. 
\end{definition}

The outcome aggregates the preference ratings and counts how often one system is preferred over the other. For simplicity, we denote $\bm{d}$ as the outcome computed using the human ratings, and $\hat{\bm{d}}$ as the outcome according to the metric. 

\section{Measuring Favoritism}
In this section, we define favoritism and derive a score to quantify it. Our view of favoritism is fundamentally relative, since it is based on comparing the behaviour of a metric with respect to the ground-truth provided by humans. In theory, one could compute favoritism compared to some other set of ground-truth labels (e.g., comparing two sets of human ratings). However, to keep our discussion simple, we treat human ratings as the unbiased ground truth, and leave deviations of this assumption to future work. 

Our measure of favoritism, called Favi-Score, is based on two observations about errors of an automated rating with respect to the human rating. First, every error is necessarily in favor of a specific system. For example, changing a $+$ to a $=$, is favoring $\pi_2$. Second, not every error has the same severity. Switching a preference from one system to another (i.e.,  $+$ to a $-$) is worse than classifying a draw as a preference (i.e., $=$ to a $-$), since the former has a larger impact on the outcome margin than the latter. This means that the favoritism measure should depend both on the proportion of errors in favor of one system and the severity of those errors.

\begin{figure}[ht!]

\begin{subfigure}{0.45\columnwidth}
\centering
\[
\begin{pmatrix}
    100 & \textcolor{cyan}{\bm{0}} & \textcolor{blue}{\bm{0}}  \\
    \textcolor{magenta}{\bm{0}}& 100 & \textcolor{cyan}{\bm{0}}  \\
    \addsum{\textcolor{red}{\bm{10}} \\\hline 110} & \addsum{\textcolor{magenta}{\bm{0}} \\\hline 100} &\addsum{ 90\\\hline 90} \\
\end{pmatrix}
\]
\caption{$\bm{C_{1}}$, $\hat{\bm{d_{1}}}$}\label{fig:example_C1}
\end{subfigure}%
\begin{subfigure}{0.45\columnwidth}
\centering
\[
\begin{pmatrix}
    100 & \textcolor{cyan}{\bm{0}} & \textcolor{blue}{\bm{0}}  \\
    \textcolor{magenta}{\bm{0}}& 100 & \textcolor{cyan}{\bm{0}}  \\
    \addsum{\textcolor{red}{\bm{0}}\\\hline 100} & \addsum{ \textcolor{magenta}{\bm{10}} \\\hline 110}&\addsum{  90 \\\hline 90}\\
\end{pmatrix}
\]  
\caption{$\bm{C_{2}}$, $\hat{\bm{d_{2}}}$}\label{fig:example_C2}
\end{subfigure}

\begin{subfigure}{0.45\columnwidth}
\centering
\[
\begin{pmatrix}
    90 & \textcolor{cyan}{\bm{0}} & \textcolor{blue}{\bm{10}}  \\
    \textcolor{magenta}{\bm{0}}& 100 & \textcolor{cyan}{\bm{0}}  \\
     \addsum{\textcolor{red}{\bm{10}}\\\hline 100} &  \addsum{\textcolor{magenta}{\bm{0}} \\\hline 100}&  \addsum{90 \\\hline 100}\\
\end{pmatrix}
\]  
\caption{$\bm{C_{3}}$, $\hat{\bm{d_{3}}}$}\label{fig:example_C3}
\end{subfigure}%
\begin{subfigure}{0.45\columnwidth}
\centering
\[
\begin{pmatrix}
    90 & \textcolor{cyan}{\bm{10}} & \textcolor{blue}{\bm{0}}  \\
    \textcolor{magenta}{\bm{0}}& 100 & \textcolor{cyan}{\bm{0}}  \\
     \addsum{\textcolor{red}{\bm{10}}\\\hline 100} &  \addsum{\textcolor{magenta}{\bm{0}} \\\hline 110}&  \addsum{90\\\hline 90} \\
\end{pmatrix}
\]  
\caption{$\bm{C_{4}}$, $\hat{\bm{d_{4}}}$}\label{fig:example_C4}
\end{subfigure}

\caption{Example Confusion Matrices $\bm{C}$ and outcomes $\hat{\bm{d}}$ according to automated metrics for the human outcome $\bm{d} = (100, 100, 100)$.}\label{fig:example_confusions}
\end{figure}

To better illustrate the interplay of these criteria, let us consider the confusion matrices depicted in Figure~\ref{fig:example_confusions} for an outcome of $\bm{d} = (100, 100, 100)$ alongside the corresponding outcome according to the metric $\hat{\bm{d}}$. We can see the error counts in the off-diagonal entries of $\bm{C_i}$, where the lower triangular sub-matrix, indicated in \textcolor{magenta}{magenta} and \textcolor{red}{red}, corresponds to errors in favor of $\pi_1$ and the upper triangular sub-matrix, indicated in \textcolor{cyan}{cyan} and \textcolor{blue}{blue}, corresponds to errors in favor of $\pi_2$. Let us first compare $\bm{C_1}$ and $\bm{C_2}$. In both cases there are $10$ errors that are all in favor of $\pi_1$. Nevertheless, $\bm{C_1}$ shows more favoritism since its errors are more severe, favoring $\pi_1$ more strongly. The consequence is visible in the outcome margin, where $mar(\hat{d}_{1}) = 110 - 90 = 20$ and $mar(\hat{d}_{2}) = 100 - 90 = 10$. $\bm{C_3}$ has more total errors, namely $20$, but they are equally distributed in favor of each system and of equal severity. Therefore, $\bm{C_3}$ does not favor either system, which is also visible in the outcome margin $mar(\hat{d}_{3}) = 0$. Finally, $\bm{C_4}$ also has $20$ errors distributed equally in favor of either system. But since the errors in favor of $\pi_1$ are more severe (i.e., mistaking + and -) than the ones in favor of $\pi_2$ (i.e., mistaking + and =), $\bm{C_4}$ favors $\pi_1$ overall. Thus, favoritism depends on how the errors are distributed, and how the severity of the mistakes are counted. The severity can be quantified in terms of how much they change the outcome according to the metric. Mistaking $+$ for $-$ leads to a change of 2 in the outcome margin, whereas mistaking $+$ with $=$ changes the outcome margin by 1. Thus, we weigh these mistakes accordingly and define the error cost: 

\begin{definition}[Directed Error Cost]
Let the following matrix be the error cost matrix:
    \[
        \bm{W} = 
        \begin{pmatrix}
        0 & -1 & -2 \\
        1 &  0 & -1 \\
        2 &  1 &  0 \\
        \end{pmatrix}
    \], where the magnitude of each entry denotes the cost for each mistake, and the sign denotes the direction of the favoritism. 
\end{definition}

Thus the cost of an error reflects its impact on the outcome margin, and the errors in favor of $\pi_1$ have a positive weight and errors in favor of $\pi_2$ have a negative weight. In cases where $m = n$ the cost $W_{mn} = 0$. Based on these observations, we will define a measure of favoritism, called Favi-Score $\Phi$, that corresponds to the expected weight of the errors of an automated rating with respect to the human rating.

\begin{definition}[Favi-Score]\label{def:favi-score}
    Given an evaluation setting $\mathcal{E}$ and its confusion matrix $C$ the Favi-Score is
    \[
    \Phi(C) = \sum_{m,n}W_{mn}\frac{C_{mn}}{E}
    \]
\end{definition}

The final score ranges from -2 to 2, where positive values indicate favoritism of $\pi_1$ and negative values indicate favoritism of $\pi_2$.

The term $\frac{C_{mn}}{E}$ corresponds to the probability of a given type of error in cases where $m \ne n$, meaning $\frac{C_{mn}}{E} = P(r^{H}[i] = m \land r^{A}[i] = n|r^{A}[i] \ne r^{H}[i])$.

Thus, this brings us to the following interpretation of the Favi-Score:
\begin{displayquote}
The Favi-Score is the expected Error Cost.
\end{displayquote}

\begin{figure}
    \centering
    \[
    \bm{C_5} = 
    \begin{pmatrix}
        360 & 180 & 60  \\
        20  & 40  & 40  \\
        \addsum{90 \\ \hline 470}  & \addsum{90\\ \hline 310}  &  \addsum{120\\ \hline 220} \\
    \end{pmatrix}
    \]
    \caption{Another example confusion matrix for illustration for $\bm{d} = (600, 100, 300)$.}
    \label{fig:example_confusion_2}
\end{figure}

There is an equivalent definition of the Favi-Score based on the outcome margins. 
\begin{lemma}\label{lemm:eq_of_proof}
    The following equality holds:
    \begin{equation}
        \Phi(C) = \frac{mar(\hat{\bm{d}}) - mar(\bm{d})}{E} 
    \end{equation}
\end{lemma}

Thus, the Favi-Score can also be interpreted as the average change in outcome margin per error.~\footnote{The proof is in Appendix~\ref{app:proofs_1}.} 

Coming back to our examples, the Favi-Scores for the confusion matrices in Figure~\ref{fig:example_confusions} are therefore: $\Phi(\bm{C_1}) = 2$, $\Phi(\bm{C_2}) = 1$, $\Phi(\bm{C_3}) = 0$, and $\Phi(\bm{C_4}) = \frac{1}{2}$. In Figure~\ref{fig:example_confusion_2} we show a more complex example. In this case more errors are in favor of $\pi_2$ ($280$ vs. $200$), but more of the severe errors are in favor of $\pi_1$ ($90$ vs. $60$). The overall Favi-Score $\Phi(\bm{C_5}) = -0.104$ shows slight favor for $\pi_2$. The score can be interpreted as each error contributing around $0.104$ increase in margin in favor of $\pi_2$.

\section{Favi-Score vs. Sign Accuracy}
In this section we explore the relation of the Favi-Score to sign accuracy. We show that the Favi-Score and sign accuracies measure different properties of a metric, and that the scores are complementary to each other.

First, we show the relationship between the Favi-Score and the system level sign accuracy. For this, we first define the system-level sign-accuracy using our notation.

\begin{definition}[\textit{System}-level sign accuracy]
Given $N = {|\Pi|\choose 2}$ different pairs of systems. The formula for the \textit{system}-level sign accuracy is given by:
\begin{equation}
\small
\frac{1}{N}\sum_{\pi^a, \pi^b} \mathbb{I}[sgn(d_+ - d_-) = sgn(\hat{d}_+ - \hat{d}_-)]
\end{equation}
\end{definition}

That is, the system level sign accuracy measures the number of times where the outcomes of the metric and the human evaluation agree on which system had more outputs of superior quality. The relationship between the  Favi-Score and the \textit{System}-level sign accuracy is that if there is no favoritism, the sign accuracy will be 1.

\begin{lemma}[No System Level Sign Change] \label{lemm:sys_level_sgn}
If $\Phi(C) = 0$, that is, there is no favoritism, then the system-level sign will remain the same under the metric. That is: $\mathbb{I}[sgn(d_+ - d_-) = sgn(\hat{d}_+ - \hat{d}_-)] = 1$.
\end{lemma}

Thus, a metric with no favoritism ensures that the outcome will lead to the same ranking. However, note that as discussed in Section~\ref{sec:intro}, the converse might not hold, i.e., a system level sign accuracy of 1 does not imply that there is no favoritism. This brings us to the following insight:
\begin{displayquote}
    Metrics with a low Favi-Score will tend to preserve the correct ranking regardless of their sign-accuracy. 
\end{displayquote}


The difference between the Favi-Score and the \textit{sample}-level sign accuracy is apparent by considering its definition: 
\begin{definition}[\textit{Sample}-level sign accuracy]
The sample-level sign accuracy is the fraction of samples, which are correctly predicted, i.e., the sum of the diagonal of $\bm{C}$ divided by the number of samples.
\begin{equation}
{\small
\begin{aligned}
&\frac{1}{|\mathcal{T}|}\sum_{m \in \{+,=,-\}} C_{m,m} = \frac{|\mathcal{T}| - E}{|\mathcal{T}|}
\end{aligned}
}
\end{equation}
\end{definition}

Thus, sign accuracy is a formula based on the diagonal entries of $C$, i.e., the fraction of samples where the metric and humans agree. Favoritism, on the other hand, is formulated via the difference of error types in favor of one system or the other, that is the upper and lower triangular sub-matrices. To illustrate this, we come back to our example in Figure~\ref{fig:example_confusion_2}, where $\Phi(C) = -0.104$, and the sample-level sign accuracy is $0.52$. A different way to showcase the difference is by assuming a fixed number of errors E, then the sign accuracy is 0 for $|\mathcal{T}| = E$, and approaches 1 for $|\mathcal{T}| \rightarrow \infty$, however, $|\mathcal{T}|$ does not influence $\Phi(C)$.
Thus, sign accuracy tells us how often we can expect the human and metric to agree on a single sample, whereas favoritism tells us in which direction the mistakes are skewed and by how much.

\begin{figure*}[!ht]
    \small
    \centering 
    \includegraphics[width=0.8\textwidth]{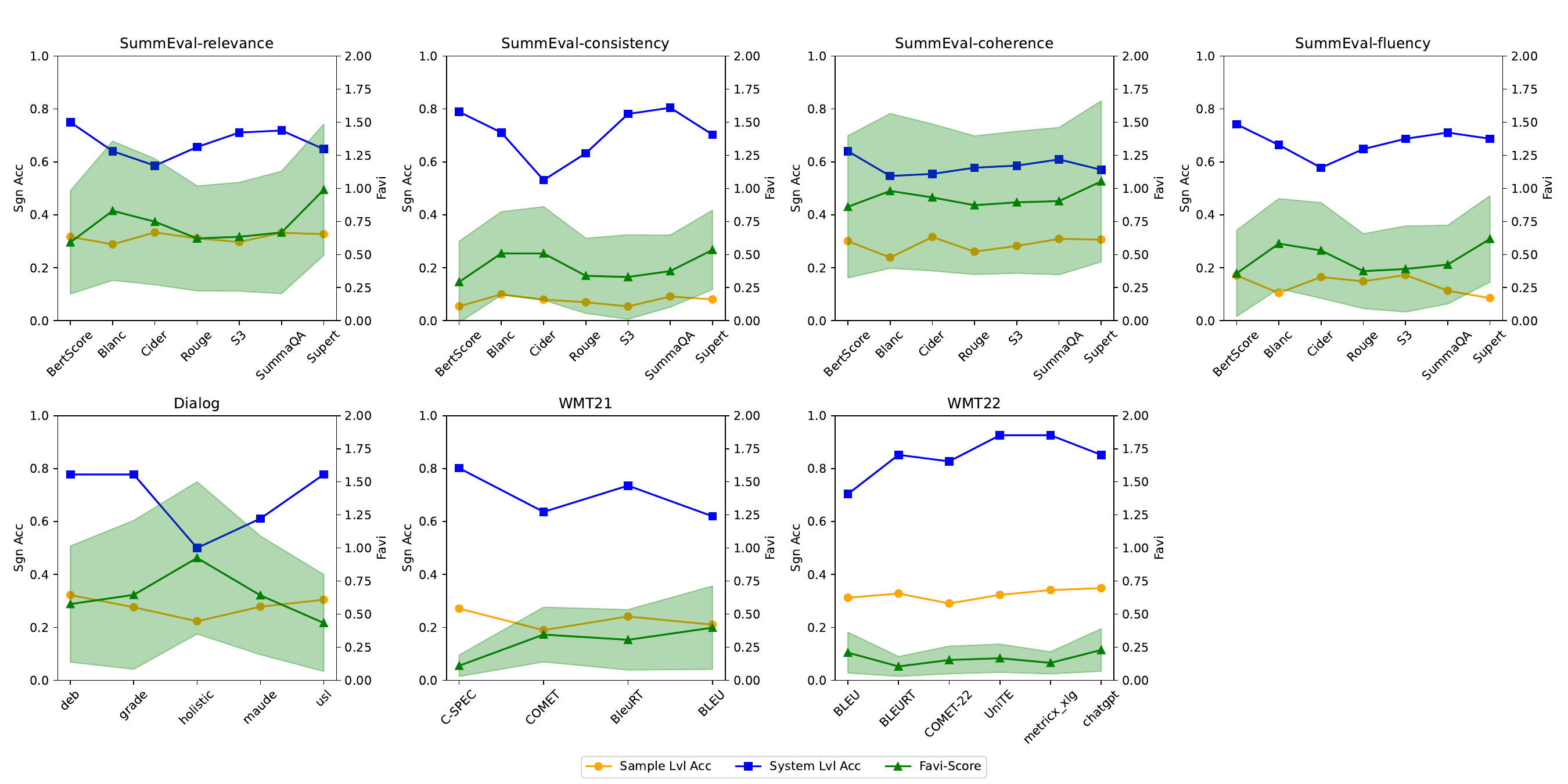}
     \caption{Visualization of the sample-level sign accuracy (orange), system-level sign accuracy (blue), and the average favi-score with a standard deviation (green).}
     \label{fig:box_plots}
\end{figure*}

\begin{figure}[!t]
\small
\centering
 \begin{subfigure}[c]{0.45\textwidth}
 \centering
     \includegraphics[width=0.9\textwidth]{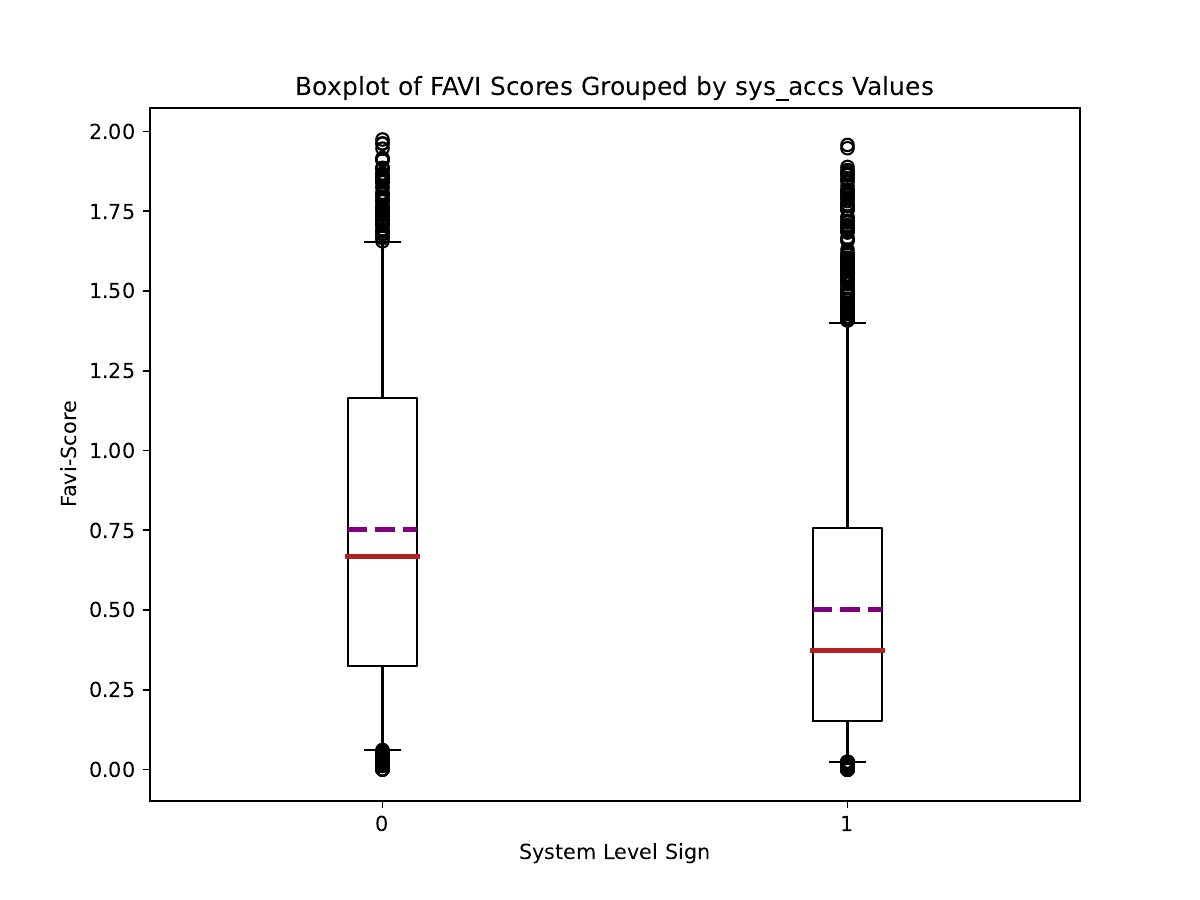}
     \caption{Favi vs System Level Sign Acc}
     \label{fig:box_all}
 \end{subfigure}
  \begin{subfigure}[c]{0.45\textwidth}
 \centering
     \includegraphics[width=0.9\textwidth]{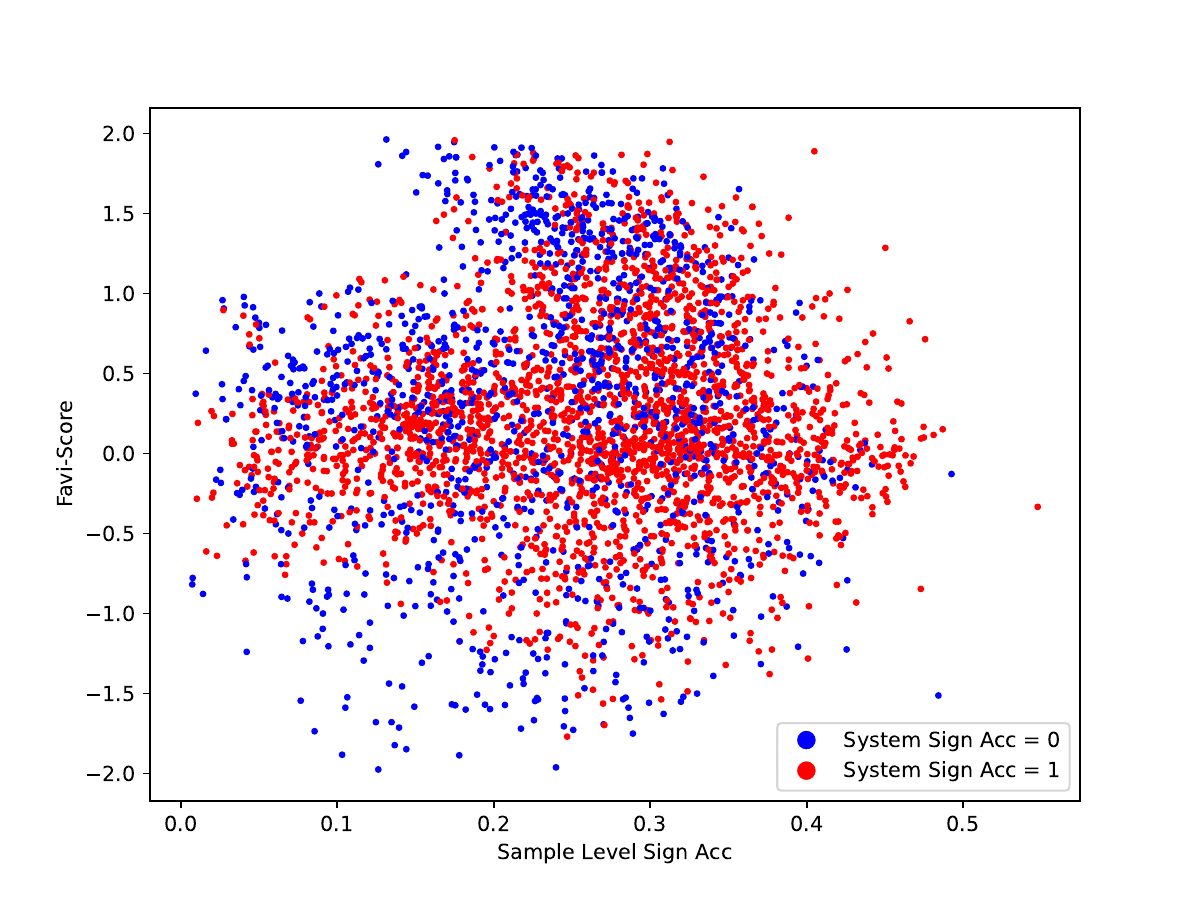}
     \caption{Favi vs Sample Level Sign Acc}
     \label{fig:scatter_all}
 \end{subfigure}
\caption{The relationship between the Favi-score and both types of sign accuracy. In the upper plot, the boxplot show the distribution of absolute Favi-scores for the cases where the System Level sign agrees for a pair (1) and for the cases where it disagrees (0). The lower plot scatters the Favi-scores vs. the sample level accuracy, where the color showcases the system-level accuracy.}
\label{fig:favi-vs-sgn}
\end{figure}



\section{Application of the Favi Score}
For the empirical application, we require data in the form highlighted by the evaluation setting $\mathcal{E} = \{i, \pi_1(i), \pi_2(i), r^H[i], r^A[i] | i \in \mathcal{T} \}$. That is where we have a set of text generation systems with paired ratings from humans and automated metrics. For this, we use the data collected by~\citet{deriu-etal-2023-correction}, which consists of three different tasks (machine translation, summarization, and dialogue systems). We extended their dataset with data from the WMT-22 metrics shared task (see Appendix~\ref{sec:exp_steup} for details). Here we provide an overview of the data (see Table~\ref{tbl:data}). Note that for most metrics and human ratings there are either scalar or Likert outputs provided, which we transform to preference ratings by comparing the values (Appendix~\ref{sec:exp_steup}).  \\
\noindent{\bf Chatbot.} The chatbot domain consists of the outputs of 5 different chatbots systems and one human reference automatically rated by 5 different metrics for the BlendedSkillTask (BST) dataset~\cite{smith-etal-2020-BST}. The metrics consist of: DEB~\cite{sai2020deb}, GRADE~\cite{huang2020-grade}, HolisticEval~\cite{pang2020towards}, MAUDE~\cite{sinha2020maude}, and USL-H~\cite{phy2020deconstruct}. There are 50 human ratings for each pair of systems, yielding $|\mathcal{E}| = 50$ paired ratings. \\
\noindent{\bf Summarization.} The summarization domain is based on the data of the SummEval framework~\cite{fabbri2021summeval}, which provides the outputs of 16 different summarization tools on the CNN/DailyMail corpus~\cite{nallapati2016abstractive} rated by 7 different automated metrics: BertScore~\cite{zhang2019bertscore}, BLANC~\cite{vasilyev2020blanc}, CIDEr~\cite{Vedantam2015cider}, Rouge-L~\cite{lin2004rouge}, S3~\cite{peyrard2017s3}, SummaQA~\cite{scialom2019summaqa}, and SUPERT~\cite{gao2020supert}. For 100 outputs, there are expert Likert ratings for each of the features: relevance, coherence, consistency, and fluency, thus $|\mathcal{E}| = 100$. \\
\noindent{\bf Machine Translation.} For machine translation, we use the WMT-21 metrics task data~\cite{freitag2021wmt21metrics}, and the WMT-22 metrics task data~\cite{freitag-etal-2022-wmt22-results}. For WMT-21 we use the English to German language pair and the news domain where eight machine translation systems were evaluated, plus three human references. We used the four most prominent metrics: BleuRT~\cite{sellam2020bleurt}, COMET~\cite{rei2020comet}, C-SPEC~\cite{takahashi2021cspec}, and sentence-level BLEU~\cite{papineni2002bleu}. There are 500 ratings by expert translators for each of the translation systems, thus, $|\mathcal{E}| = 500$. For WMT-22, we use the  English to German language pair and selected eight machine translation systems plus one human reference, and five metrics. Namely, BLEU, BleuRT, COMET22~\cite{rei-etal-2022-comet}, UniTE~\cite{wan-etal-2022-unite}, and METRICX\_XL.~\footnote{No reference available.} We additionally added \emph{ChatGPT}~\footnote{gpt-3.5-turbo-0613} as a preference metric (see Appendix~\ref{app:chatgpt}). For 1315 samples, expert ratings were provided, thus $|\mathcal{E}| = 1315$.

\begin{table}[t!]
\centering
\small
\resizebox{.5\textwidth}{!}{
\input{tables/data.tex}
}
\caption{Overview of Data Statistics. The ratings refer to the number of ratings available for each pair of TG systems.}
\label{tbl:data} 
\end{table}

\begin{figure}[!t]
\small
\centering
\includegraphics[width=0.49\textwidth]{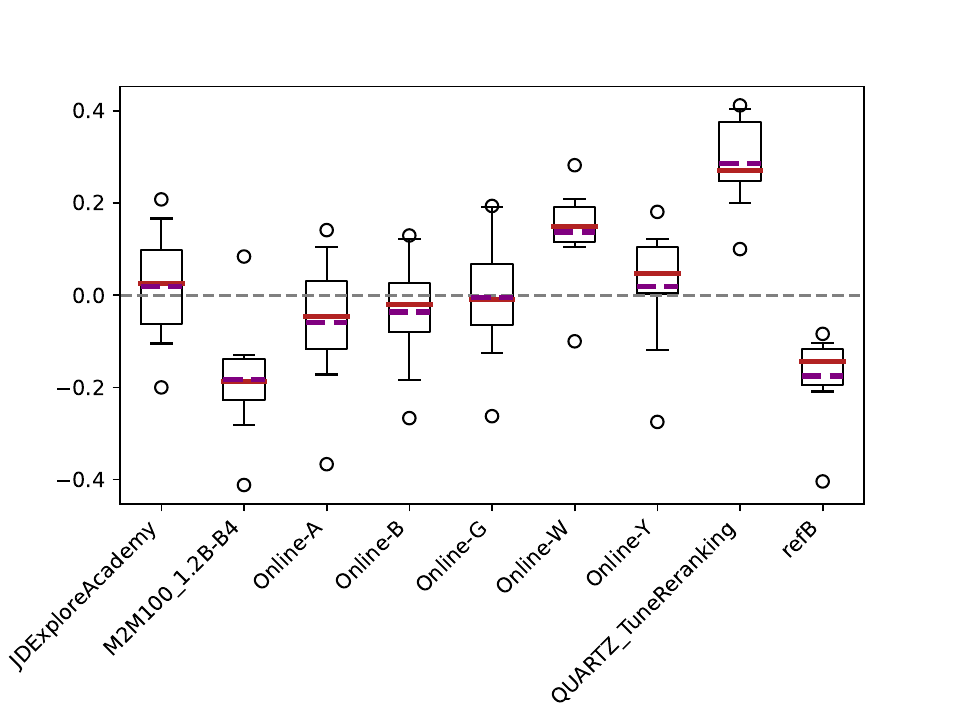}
\caption{Box Plot of the Favi-Scores for each System for the COMET-22 metric.}
\label{fig:favi-per-system}
\end{figure}

\subsection{Results \& Discussion}
\textbf{Overall Results.} We compare the Favi-Score to the sample- and system-level sign accuracy. Since the Favi-Score is computed for each pair of systems, we report the absolute average Favi-Score over each pair of systems alongside the standard-deviation, while for the two sign accuracy scores, we report the overall score. Figure~\ref{fig:box_plots} shows the three scores for each domain and metric. Note that for the Favi-Score (green) we also report the standard deviation (green area). First, we note that each metric exhibits a certain degree of favoritism, with some degree of variance with regards to the system pairs. The magnitude of the Favi-Score varies depending on the domain, for instance, machine-translation metrics exhibit a lower amount of favoritism, while the summarization metrics exhibit a larger amount as well as a larger variance. 

\textbf{System Pair Analysis.} To get a better understanding of the interplay between the three measures, we compute the sign-accuracy on the system-pair level (for the system-level sign accuracy this means that we return whether the signs agree or not). First, we investigate whether the system-level sign accuracy correlates to the Favi-Score. In Figure~\ref{fig:box_all}, we show the Favi-Scores depending on whether the sign agrees on the system level (1) or not (0). We observe that system-pairs where the sign does not agree tend to have a larger Favi-Score, which is expected according to Lemma~\ref{lemm:sys_level_sgn}. In fact, there is a low-to-moderate inverse correlation between the two scores of Spearman's-$\rho = -0.25$. The correlation of the Favi-Score to the sample-level sign accuracy is depicted in Figure~\ref{fig:scatter_all}, which shows a small correlation ( Spearman's-$\rho = -0.01$). We note that very high sample-level accuracies are associated with smaller Favi-Scores. 

\textbf{Return to Motivating Example.}  To return to the motivating example from Section~\ref{sec:mot_exp}, we showcase the Favi-Scores for each system (one-vs-all) according to COMET-22 in Figure~\ref{fig:favi-per-system}. That is we computed the Favi-Score for each pair of systems, and display for each system all Favi-Scores in terms of a box-plot. COMET-22 strongly disfavors \emph{ref-B} (i.e., the human reference), since the Favi-Score is negative for \emph{ref-B}, no matter the system we compare against.
This matches the DAG in Figure~\ref{fig:dag1} in which  \emph{ref-B} is ranked significantly lower than in the human ranking. The \emph{M2M100} system is also strongly disfavored. However, this is not visible from the DAGs alone, where it is ranked last, both according to COMET-22 and human judgements. This is a case where the Favi-Score uncovers a potentially undesired outcome that could go unnoticed otherwise, highlighting the utility of the Favi-Score as a diagnostic tool. Finally, we observe a strong favoritism with regards to the~\emph{QUARTZ} system, which is ranked significantly higher according to COMET-22. Thus, the Favi-Score is also useful as a diagnostic tool and provides an explanation for observed errors in system rankings. See Appendix~\ref{app:examples} for more such examples. 

The results of the application of the Favi-Score to evaluate metrics shows that it measures a different aspect of the metrics behaviour than the sign-accuracy scores. Thus, we advocate the usage of the Favi-Score alongside existing scores when a novel metric for generative systems is developed.

\section{Conclusion}
We motivated and defined the Favi-Score, a simple-to-implement and easy to interpret score, which measures how much an automated preference metric favors the outputs of a generative AI system. Our investigations show that the Favi-Score differs from the sign accuracy scores, and is to be used complementary to them. The empirical application of the Favi-Score to real-world setting reveals that all metrics exhibit some degree of favoritism, which needs to be accounted for when applying these metrics, to better interpret results. We envision that the Favi-Score is used as a diagnostic tool to understand which systems are favored by a metric. 

\section*{Acknowledgments}
This work was supported by the Swiss Innovation Agency (Innosuisse) within the project "Unified Model for Evaluation of Text Generation Systems (UniVal)" [200020\_219819].

\section*{Limitations}

\noindent\textbf{Relative Measure.} Our definition of favoritism is fundamentally relative, meaning we can only say that some automated preference ratings display favoritism with respect to a specific set of human ratings. \\
\noindent\textbf{Anchoring.} This makes it crucial to prudently select the reference ratings that are used for the comparison. In this work, we treat human ratings as the gold standard reference despite their limitations~\citep{amidei-etal-2018-rethinking, belz2021reprogen}. \\
\noindent\textbf{Comparisons.} The Favi-Score is defined for a given evaluation setting $\mathcal{E}$, meaning for a fixed set of automated ratings, human ratings, and system pair. If we replace the the automated ratings, the resulting new Favi-Score is directly comparable. On the other hand, comparing the Favi-Score for one system pair to another, even for the same automated metric, means we have to assume a degree of consistency between reference ratings. \\
\noindent\textbf{Alternative Definitions.} Our definition of favoritism is narrowly defined based on (perceived) errors with respect to human ratings. In particular, we do not consider the content of either the inputs or outputs of various systems under test. Even within our narrow framework, one could imagine alternative definitions taking into account how a change in outcome affects significance tests.\\
\noindent\textbf{Robustness.} Throughout this work, we consider confusion matrices based on reasonably sized evaluation settings. For smaller settings the utility of the Favi-Score can be limited. For example, if there is only $1$ error then the only possible values of the Favi-Score are $\pm 1$ and $\pm 2$. This could potentially be mitigated by a full probabilistic treatment of the evaluation setting, which is outside the scope of this work. \\
\noindent\textbf{Diagnostic.} While we are convinced that the Favi-Score adds a useful diagnostic to a practitioner's toolbox, it does not provide any remedies. For example, to lower the Favi-Score, one can try to distribute errors more evenly, but it is not clear how this should be achieved in a concrete case.
\bibliography{custom}

\appendix

\section{Proof of Equal Definition}
\label{app:proofs_1}
Here, we prove Lemma~\ref{lemm:eq_of_proof}, which states:

\begin{equation}
     \sum_{m,n}W_{mn}\frac{C_{mn}}{E} = \frac{mar(\hat{\bm{d}}) - mar(\bm{d})}{E} 
\end{equation}

We start the proof by expanding both sides of the equation. We start with the left side. Let for simplicity $r^H_n = \mathbb{I}[r^H[i] = n]$

\begin{align}
\label{eq:left_exp}
\begin{split}
\small
\Phi(C) &  = \sum_{m,n}W_{mn}\frac{C_{mn}}{E} \\
& = \frac{1}{E} \sum_{m,n}W_{mn}*C_{mn} \\
& = \frac{1}{E} \sum_{i \in \mathcal{T}}\sum_{m,n}W_{mn}\mathbb{I}[r^H[i] = m \land r^A[i] = n] \\
& = \frac{1}{E} \sum_{i \in \mathcal{T}} \sum_{m,n}W_{mn}\mathbb{I}[r^H[i] = m]\mathbb{I}[r^A[i] = n] \\
& = \frac{1}{E} \sum_{i \in \mathcal{T}} [-r^H_+r^A_= + -2r^H_+r^A_- -r^H_=r^A_-  \\
& + r^H_=r^A_+ + 2r^H_-r^A_+ + r^H_-r^A_=]
\end{split}
\end{align}

Let us now expand the right side:
\begin{align}
\label{eq:right_exp}
\begin{split}
\small
\Phi(C) &  = \frac{mar(\hat{\bm{d}}) - mar(\bm{d})}{E} \\
& = \frac{1}{E}((\hat{d}_+ - \hat{d}_-) - (d_+ - d_-)) \\
& = \frac{1}{E}(\sum_{i \in \mathcal{T}}r^A_+ - \sum_{i \in \mathcal{T}}r^A_- - \sum_{i \in \mathcal{T}}r^H_+ + \sum_{i \in \mathcal{T}}r^H_-) \\
& =  \frac{1}{E}\sum_{i \in \mathcal{T}}(r^A_+ - r^A_- - r^H_+ + r^H_-)
\end{split}
\end{align}

It is now apparent that we need to find an equivalence of the terms inside the sum. That is:

\begin{align}
\begin{split}
     &-r^H_+r^A_= + -2r^H_+r^A_- -r^H_=r^A_- \\
     & + r^H_=r^A_+ + 2r^H_-r^A_+ + r^H_-r^A_= \\
     &= r^A_+ - r^A_- - r^H_+ + r^H_-
\end{split}
\end{align}

We note that on the left hand side only at most one term of the sum can be non-zero. On the right hand side at most one term for the automated metric can be 1 at the same time, and one for the human rating. We complete the proof by simple enumeration of all possible values of $r^A_n$ and $r^H_m$.

\input{tables/app_proof_enumeration}

Thus, for all possible combinations of  $r^A_n$ and $r^H_m$, both sides yield the same value, thus, showing the equality of the two definitions. 

\section{Proof of No System Level Sing Change}
Here, we proof Lemma~\ref{lemm:sys_level_sgn}, i.e., that 
\begin{align}
    \begin{split}
    \small
        &\Phi(C) = 0 \\
        &\implies \mathbb{I}[sgn(d_+ - d_-), sgn(\hat{d}_+ - \hat{d}_-)] = 1
    \end{split}
\end{align}

The proof is a simple application of the alternative definition of the Favi-Score.
\begin{align}
\begin{split}
    &\Phi(C) = 0 \\
    & \iff \frac{mar(\hat{\bm{d}}) - mar(\bm{d})}{E} = 0 \\
    & \iff mar(\hat{\bm{d}}) - mar(\bm{d}) = 0 \\
    & \iff mar(\hat{\bm{d}}) - mar(\bm{d}) = 0 \\
    & \iff mar(\hat{\bm{d}}) =  mar(\bm{d}) \\
    & \iff \hat{d}_+ - \hat{d}_- =  d_+ - d_- \\
    & \implies sgn(\hat{d}_+ - \hat{d}_-) = sgn(d_+ - d_-)
\end{split}
\end{align}

Note that the last line of the proof is an implication, which is not reversible. Thus if the sign of both margins are equal, we cannot conclude that there is no favoritism. 

\section{Experimental Setup Details}
\label{sec:exp_steup}

\subsection{Convert Scalar to Preference}

Since most metrics return a scalar value, we need to transform them into a preference rating. Similarly, in many cases human ratings are available as scalar or ordinal judgements for a single input and output pair. In theses cases we assume that we have paired ratings, meaning that we have system output and rating for each test input and all systems. We can then compare the individual scalar (or ordinal) ratings to derive a preference rating, assuming that higher values indicate higher quality.

\begin{definition}[Scalar Metric]
    We call real valued functions of inputs and outputs \textbf{scalar metrics}:
    $\mathcal{M}_s: \mathcal{I} \times \mathcal{O} \rightarrow \mathbb{R}$.
\end{definition}

A preference metric can be constructed from a scalar metric as follows:
\begin{definition}
\label{def:derived_metric}
    The \textbf{derived preference metric} $\mathcal{M}$ of a given scalar metric $\mathcal{M}_{s}$
    is defined as
    \small
    \[
        \mathcal{M}(i, o_1, o_2) =
        \begin{cases}
            + & \mathcal{M}_{s}(i, o_1) > \mathcal{M}_{s}(i, o_2) \\
            = & \textit{otherwise} \\
            - & \mathcal{M}_{s}(i, o_1) < \mathcal{M}_{s}(i, o_2) \\
        \end{cases}
    \]
\end{definition}

\subsection{WMT-22 ChatGPT}\label{app:chatgpt}

Additionally to metrics participating in the WMT-22 metrics shared task~\citep{freitag-etal-2022-wmt22-results}, we include ratings from GPT3.5-Turbo (0613)~\citep{gpt35turbo0613} prompted as a preference metric~\citep{pandalm2024}. We provided both a static \emph{system prompt} describing the rating task and a \emph{user prompt} for a concrete source sentence and pair of target translations. We used the following system prompt:
\begin{displayquote}
    You act as an expert translator giving detailed feedback about candidate translations provided by users. Consider in particular the spelling, grammar, accuracy, and fluency of different translations. Make sure you give detailed information why one translation might be better than another.
\end{displayquote}
We used the following user prompt, where the placeholders \emph{source}, \emph{hyp\_a}, and \emph{hyp\_b} were replaced with a concrete source sentence and translations from systems $\pi_1$ and $\pi_2$:
\begin{displayquote}
    Please give feedback for the following translations:
    
    Original Sentence:
    \{source\}
    
    Candidate A:
    \{hyp\_a\}
    
    Candidate B:
    \{hyp\_b\}
\end{displayquote}
We used the \emph{function calling} API~\citep{openai-functioncall} to get a structured output rating and asked the model to provide its preference (\emph{Candidate A}, \emph{Candidate B}, or \emph{No Preference}), specific feedback for each candidate translation, and an explanation for its decision. For our evaluation, we discard the feedback and explanation texts which were included to elicit behavior analogous to \emph{chain of thought}~\citep{chain-of-thought} prompting. We note that we received preference ratings that were uniquely interpretable as one of the 3 options in almost all cases. For 3 items, the model responded with an empty string as preference rating, which we assigned as \emph{No Preference}.

For our experiments, we assume that the ratings are symmetric with respect to system order, meaning that if we have already collected the ratings for the pair $(\pi_1, \pi_2)$, we invert the ratings ($+$ to $-$ and vice-versa) for the pair $(\pi_2, \pi_1)$. We enumerated systems in lexicographical order.

\subsubsection{Symmetry of Preference Ratings}

In the main text we treat the ratings from GPT3.5-Turbo as symmetric. This is mainly to keep the analysis simple. We note that according to Definition~\ref{def:derived_metric} the other preference metrics are already symmetric.

\begin{figure}[ht!]
    \centering
    \includegraphics[width=\columnwidth]{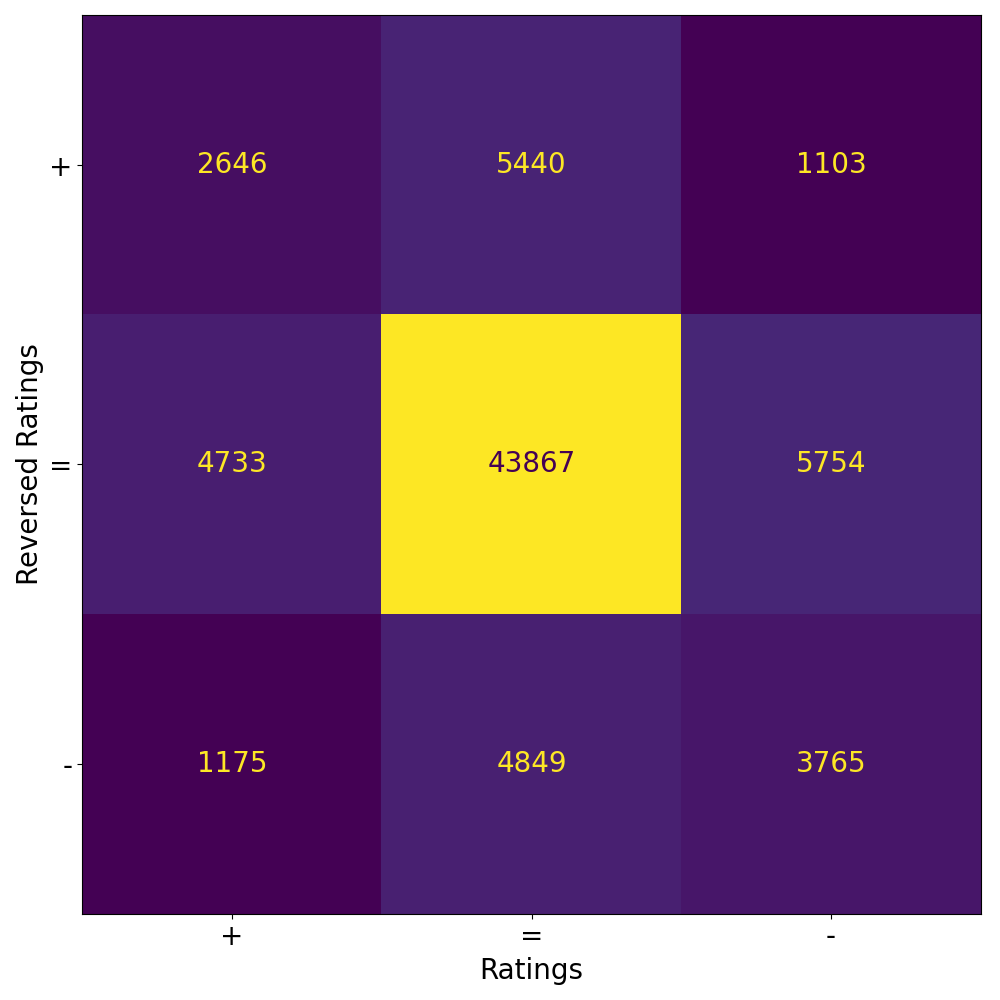}
    \caption{Confusion Matrix of reversed ratings with respect to the original GPT3.5 Turbo ratings.}
    \label{fig:gpt-confusion}
\end{figure}

We nevertheless also collected ratings for flipped system pairs and noticed discrepancies. We show the confusion matrix in Figure~\ref{fig:gpt-confusion}. The overall  accuracy is $69\%$ and the intra-rater Krippendorff-$\alpha$~\citep{krippendorff-alpha} is $0.249$.

\section{Examples of Favoritism Consequences}
\label{app:examples}
In this section, we show more examples of DAGs with the system-wise Favi-Score box plots.

\subsection{Dialogue}

\begin{figure*}[t!]
\small
\centering
 \begin{subfigure}[c]{0.32\textwidth}
 \centering
     \includegraphics[width=1\textwidth]{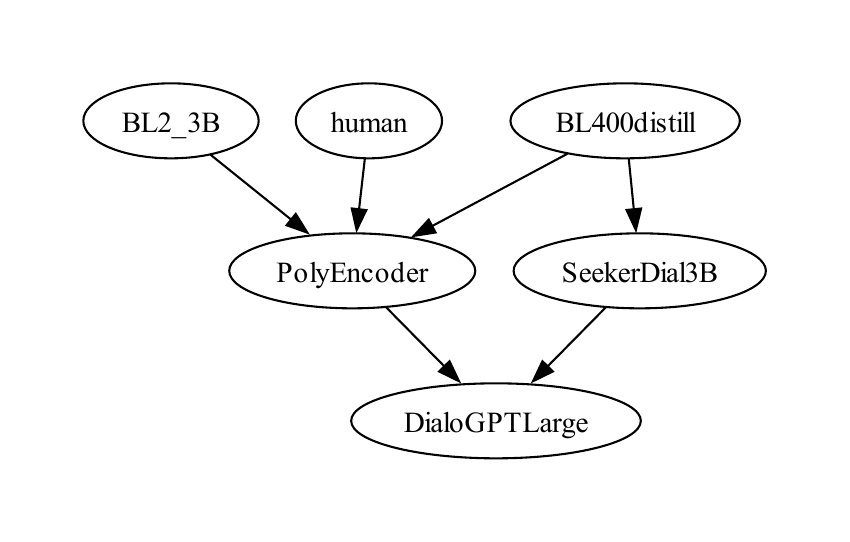}
     \caption{Human}
     \label{fig:ex_dag_dialog_deb_hum}
 \end{subfigure}
\begin{subfigure}[c]{0.32\textwidth}
 \centering
     \includegraphics[width=1\textwidth]{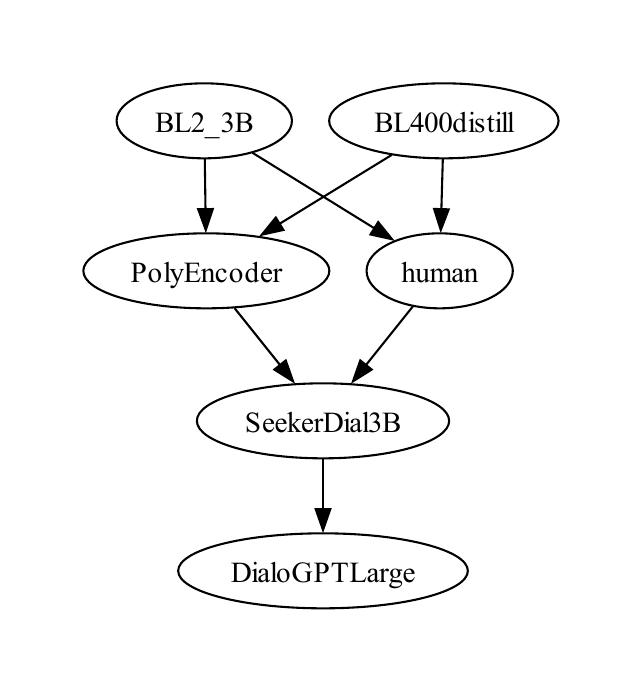}
     \caption{DEB}
     \label{fig:ex_dag_dialog_deb_deb}
 \end{subfigure}
 \begin{subfigure}[c]{0.32\textwidth}
 \centering
     \includegraphics[width=1\textwidth]{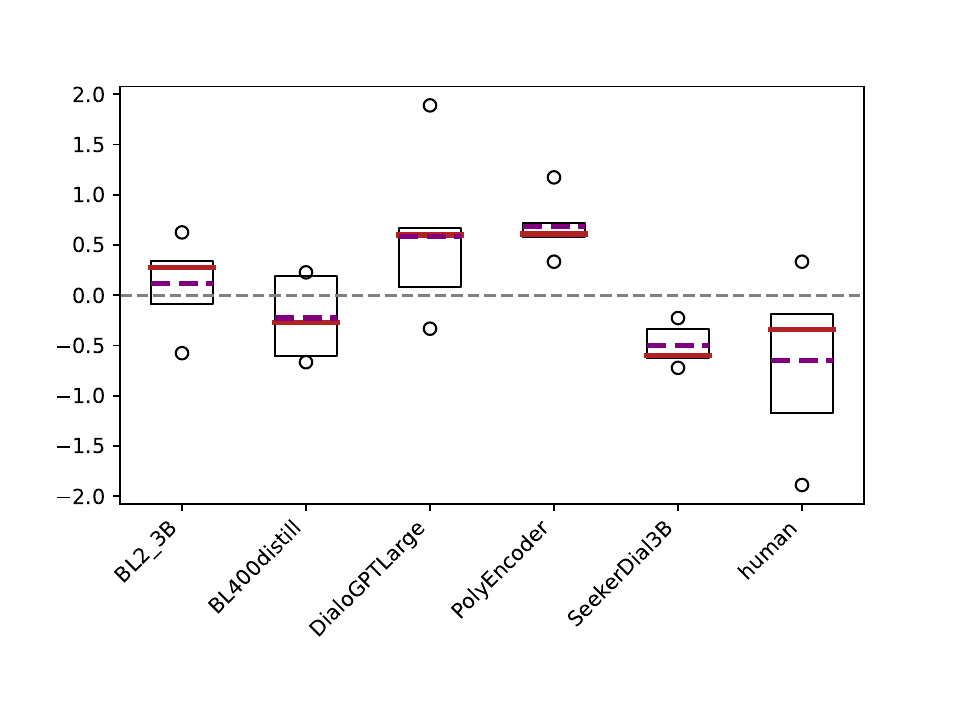}
     \caption{DEB}
     \label{fig:ex_dag_dialog_deb_box}
 \end{subfigure}
\caption{ Visualization of the consequences of unfair metrics by comparing the human ranking to the DEB ranking. The ranking of a set of systems is depicted as a Directed Acyclic Graph, where an edge from system A to system B states that system A "wins" against system B. Here, a win is determined by a sign test~\cite{sign_tests} at a 95\% confidence threshold.}
\label{fig:ex_dag_dialog_deb}
\end{figure*}

The DAG according to human ratings is depicted in Figure~\ref{fig:ex_dag_dialog_deb_hum}. The human reference is on the top row, while \emph{DialoGPTLarge} is rated as the worst system according to the human ratings. According to the DEB Metric~\cite{sai2020deb}, human responses are rated much lower than according to the human raters, while the \emph{PolyEncoder} is rated better than the \emph{SeekerDIal3B}. This matches the Favi-Scores depicted in Figure~\ref{fig:ex_dag_dialog_deb_box}, where the human reference is clearly disfavored, and the \emph{PolyEncoder} has an overall positive Favi-Score. The \emph{DialoGPTLarge} is also favored by the DEB metric, however, it does not have an impact on the new ranking. We also note that \emph{PolyEncoder} is rated as being better than \emph{SeekerDial3B} according to the metric, which is also a consequence of favoritism, as \emph{SeekerDial3B} is heavily disfavored. In fact, the Favi-Score between the two systems lies at $0.72$ in favor of \emph{PolyEncoder}.

\begin{figure*}[t!]
\small
\centering
 \begin{subfigure}[c]{0.32\textwidth}
 \centering
     \includegraphics[width=1\textwidth]{figures/examples/dialog/human-human_only-naive.gv.pdf}
     \caption{Human}
     \label{fig:ex_maude_dialog_deb_hum}
 \end{subfigure}
\begin{subfigure}[c]{0.32\textwidth}
 \centering
     \includegraphics[width=1\textwidth]{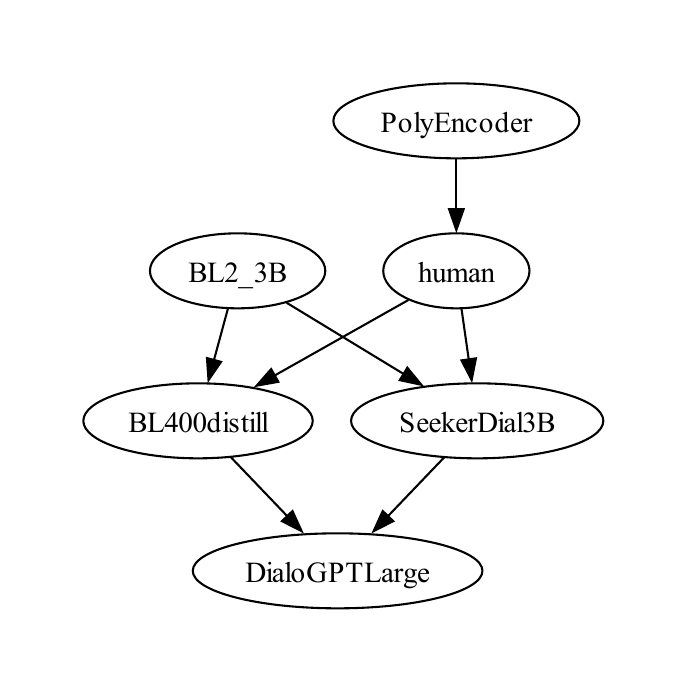}
     \caption{MAUDE}
     \label{fig:ex_maude_dialog_deb_deb}
 \end{subfigure}
 \begin{subfigure}[c]{0.32\textwidth}
 \centering
     \includegraphics[width=1\textwidth]{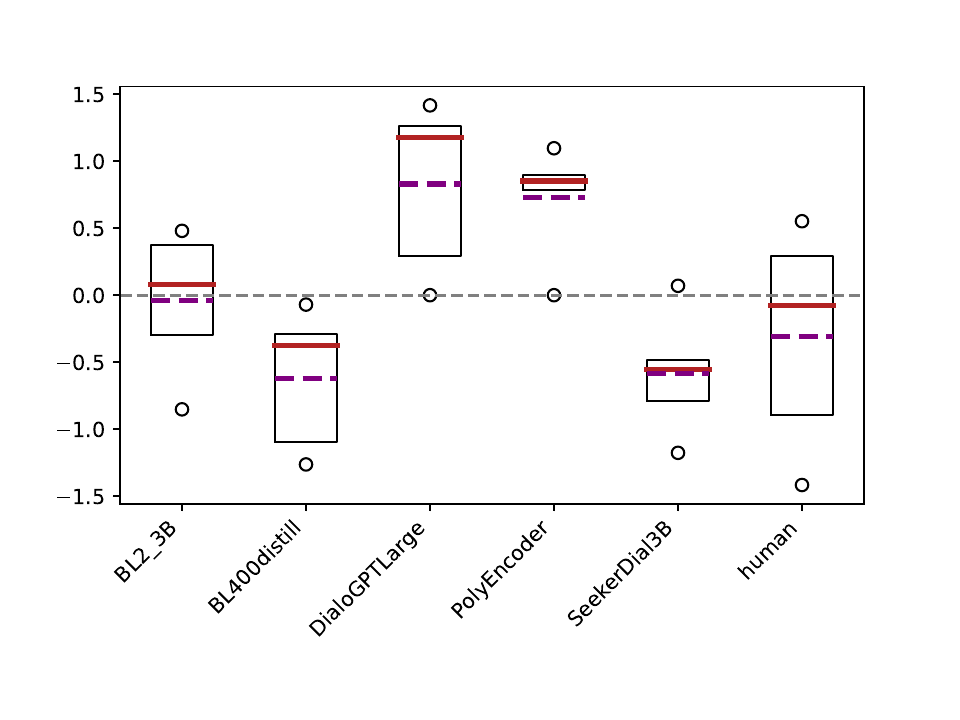}
     \caption{MAUDE}
     \label{fig:ex_maude_dialog_deb_box}
 \end{subfigure}
\caption{ Visualization of the consequences of unfair metrics by comparing the human ranking to the MAUDE ranking. The ranking of a set of systems is depicted as a Directed Acyclic Graph, where an edge from system A to system B states that system A "wins" against system B. Here, a win is determined by a sign test~\cite{sign_tests} at a 95\% confidence threshold.}
\label{fig:ex_dag_dialog_maude}
\end{figure*}

In Figure~\ref{fig:ex_dag_dialog_maude}, the analysis for the MAUDE~\cite{sinha2020maude} metric is depicted. MAUDE clearly prefers the outputs of \emph{PolyEncoder}, which is also clear from the Favi-Score. Similar to DEB, MAUDE also prefers the outputs of \emph{DialoGPTLarge}, however without changing the ranking. MAUDE has no clear disfavor against the human references like the DEB metric. Also \emph{BL2\_3B} is favored compared to \emph{BL400distill} with a Favi-Score of $0.375$ in favor. 

\subsection{Summarization}

\begin{figure*}[t!]
\small
\centering
 \begin{subfigure}[c]{0.32\textwidth}
 \centering
     \includegraphics[width=1\textwidth]{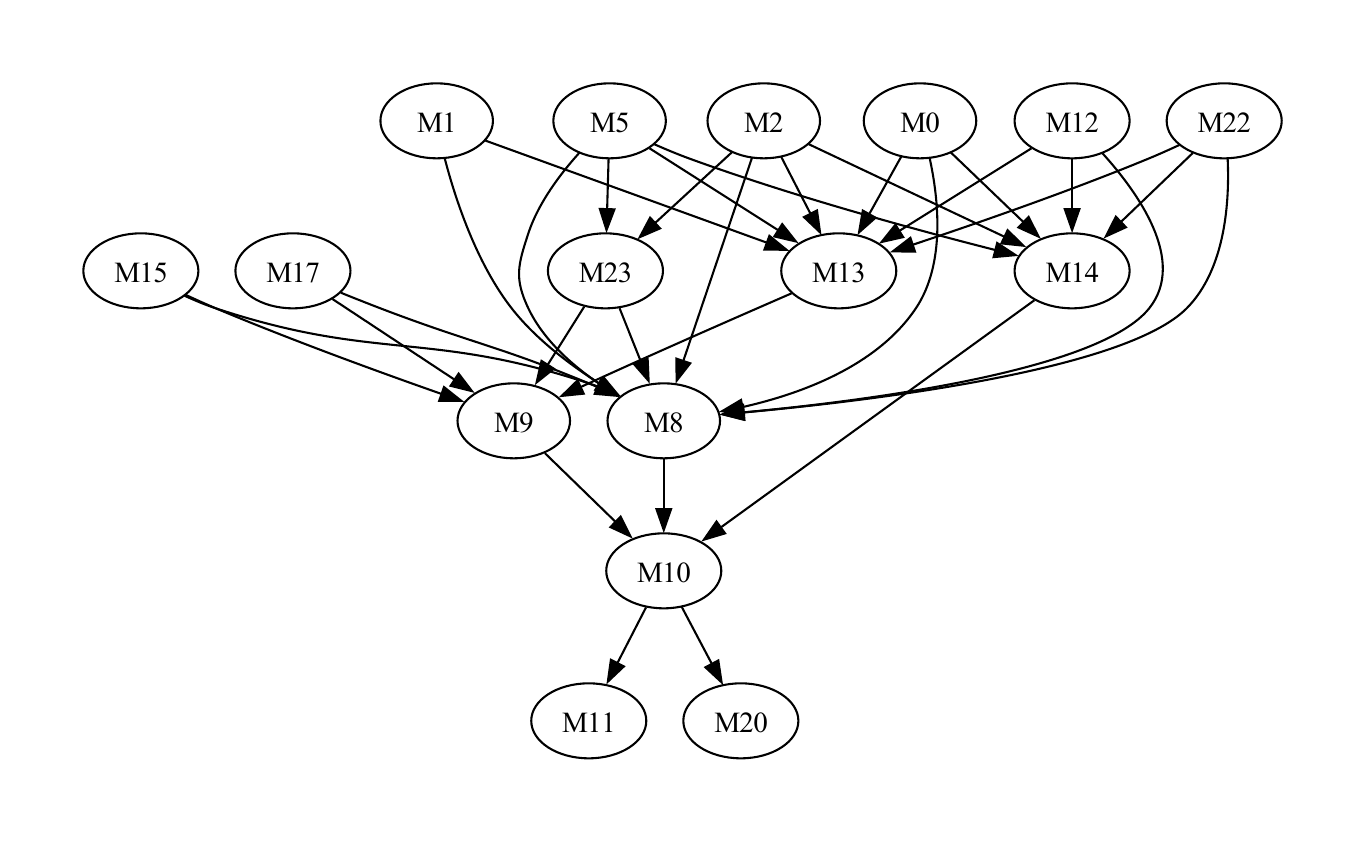}
     \caption{Human}
     \label{fig:ex_ex_bert_summ_bert_hum}
 \end{subfigure}
\begin{subfigure}[c]{0.32\textwidth}
 \centering
     \includegraphics[width=1\textwidth]{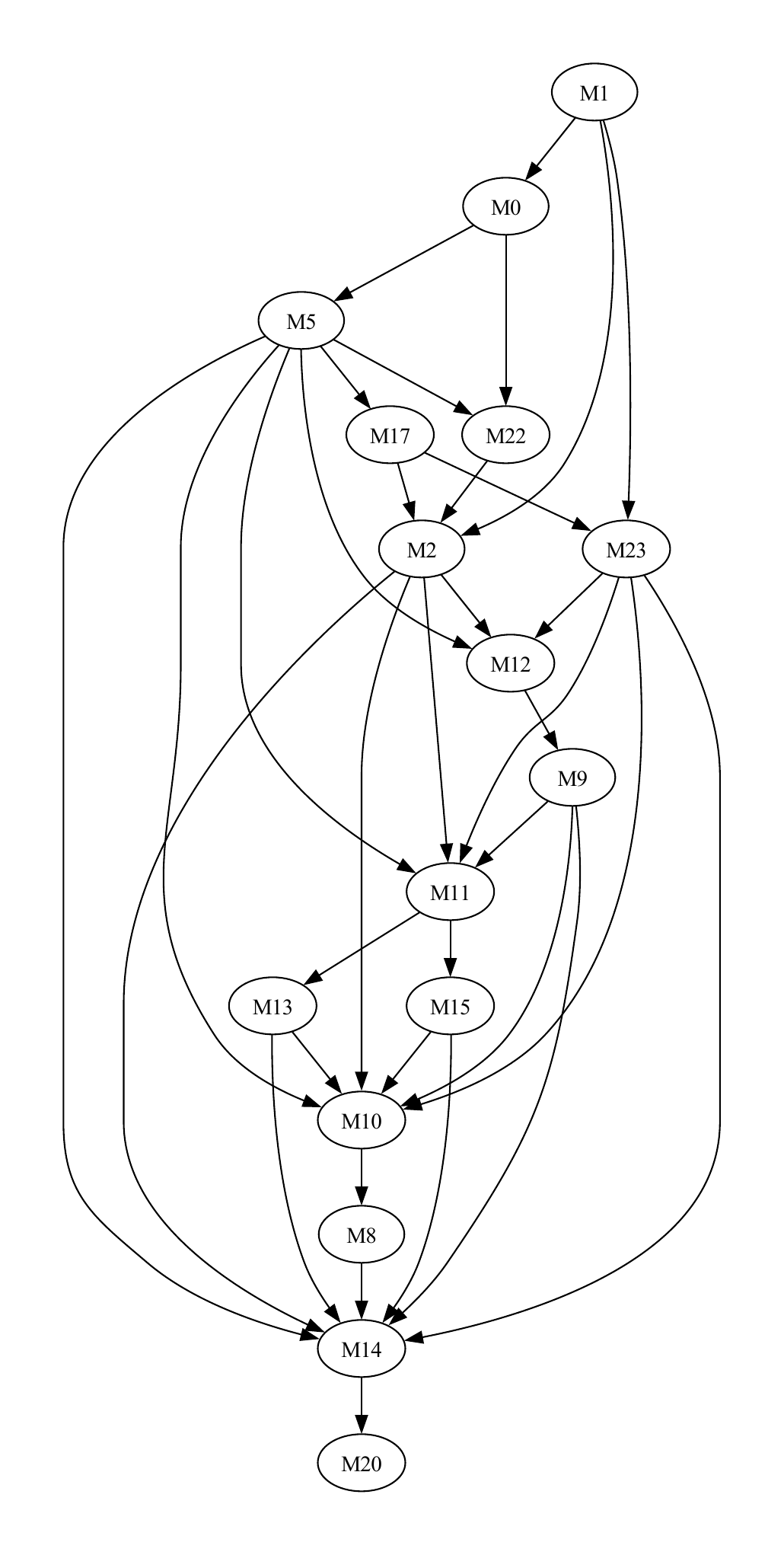}
     \caption{BertScore}
     \label{fig:ex_bert_summ_bert}
 \end{subfigure}
 \begin{subfigure}[c]{0.32\textwidth}
 \centering
     \includegraphics[width=1\textwidth]{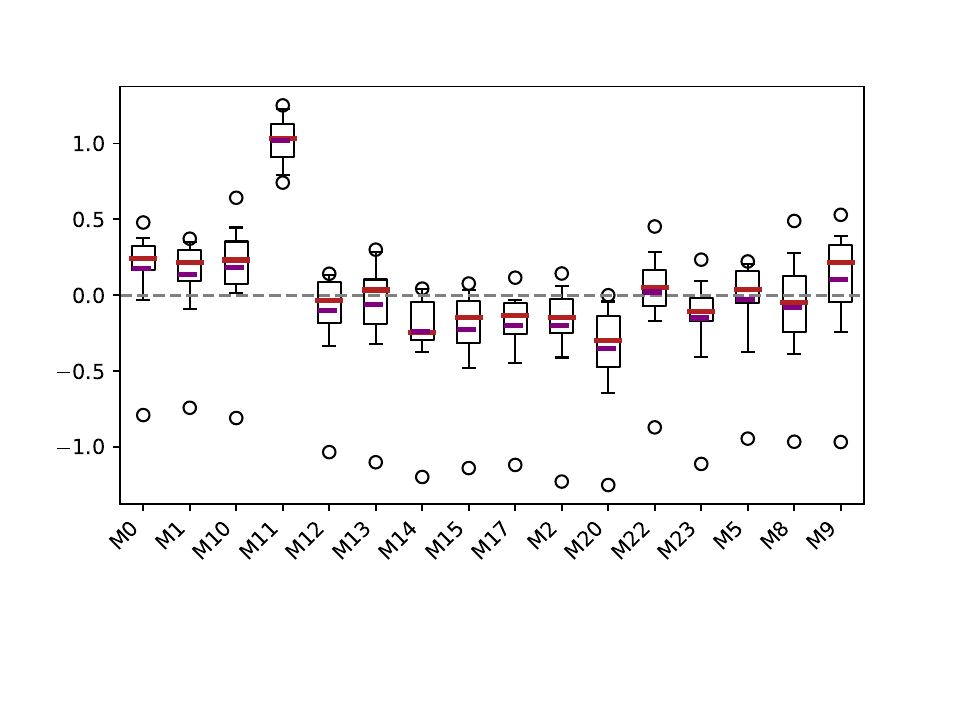}
     \caption{BertScore}
     \label{fig:ex_bert_summ_box}
 \end{subfigure}
\caption{ Visualization of the consequences of unfair metrics by comparing the human ranking to the BertScore ranking. The ranking of a set of systems is depicted as a Directed Acyclic Graph, where an edge from system A to system B states that system A "wins" against system B. Here, a win is determined by a sign test~\cite{sign_tests} at a 95\% confidence threshold.}
\label{fig:ex_dag_summ_bert}
\end{figure*}

Figure~\ref{fig:ex_dag_summ_bert} shows the analysis for BertScore~\cite{zhang2019bertscore} on the SummEval data for the Consistency feature. The strongest favoritism is shown towards \emph{M11} (a decoder focused on promoting novelty), which is the worst according to the human ranking, but is in solid mid-field according to the metric. While \emph{M20} (a GPT2 model) is strongly disfavored by the metric. We also note that BertScore has a high system-level sign accuracy (0.789), while having an exceptionally low sample level accuracy (0.05). However, since the Favi-Score is low as well (mean of 0.29), it showcases that the Favi-Score is more influential on the final ranking than the sample level accuracy. 

\begin{figure*}[t!]
\small
\centering
 \begin{subfigure}[c]{0.32\textwidth}
 \centering
     \includegraphics[width=1\textwidth]{figures/examples/summeval/human-human_only-naive.gv.pdf}
     \caption{Human}
     \label{fig:ex_rouge_summ_rouge_hum}
 \end{subfigure}
\begin{subfigure}[c]{0.32\textwidth}
 \centering
     \includegraphics[width=1\textwidth]{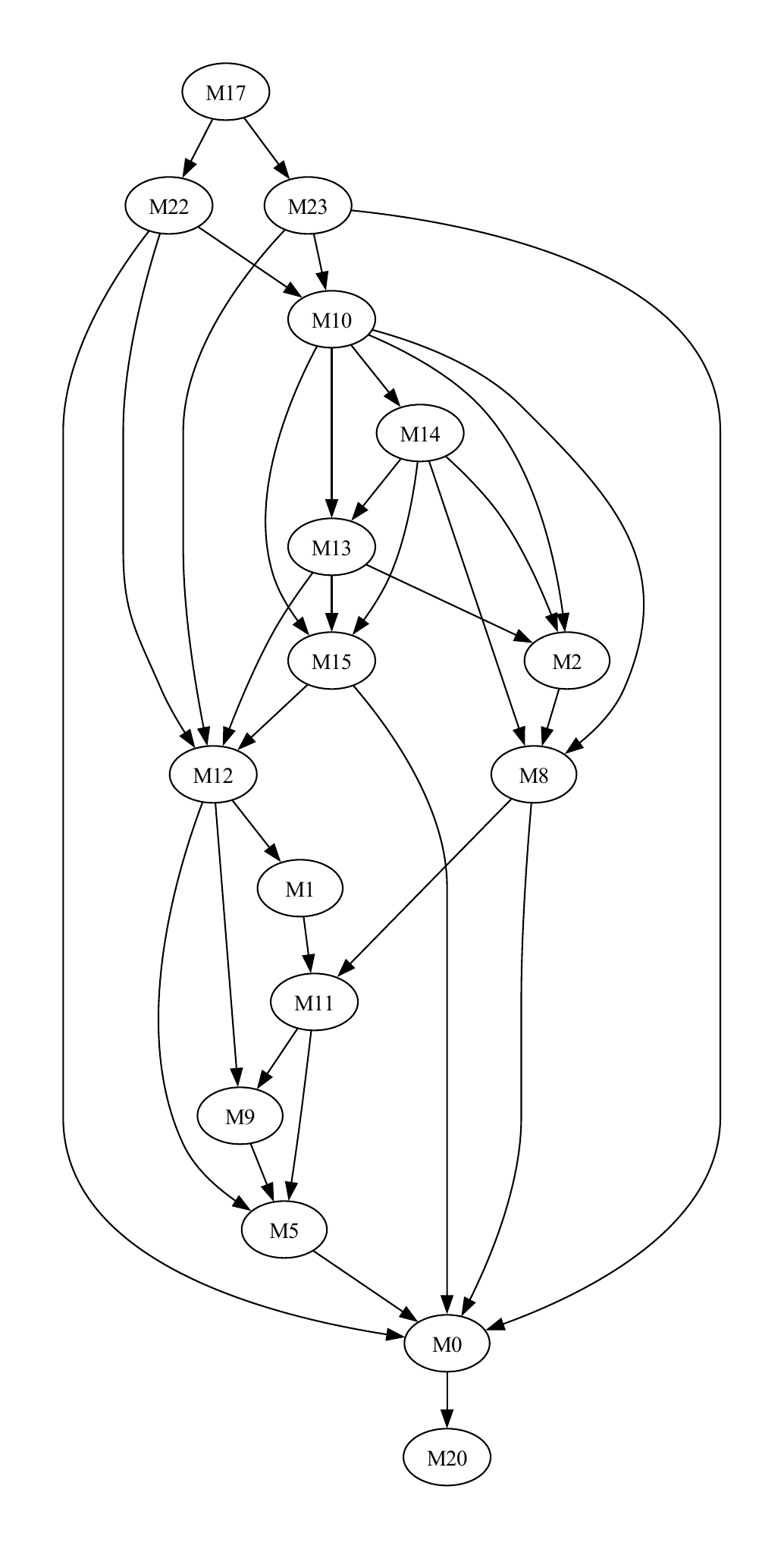}
     \caption{ROUGE}
     \label{fig:ex_rouge_summrouge}
 \end{subfigure}
 \begin{subfigure}[c]{0.32\textwidth}
 \centering
     \includegraphics[width=1\textwidth]{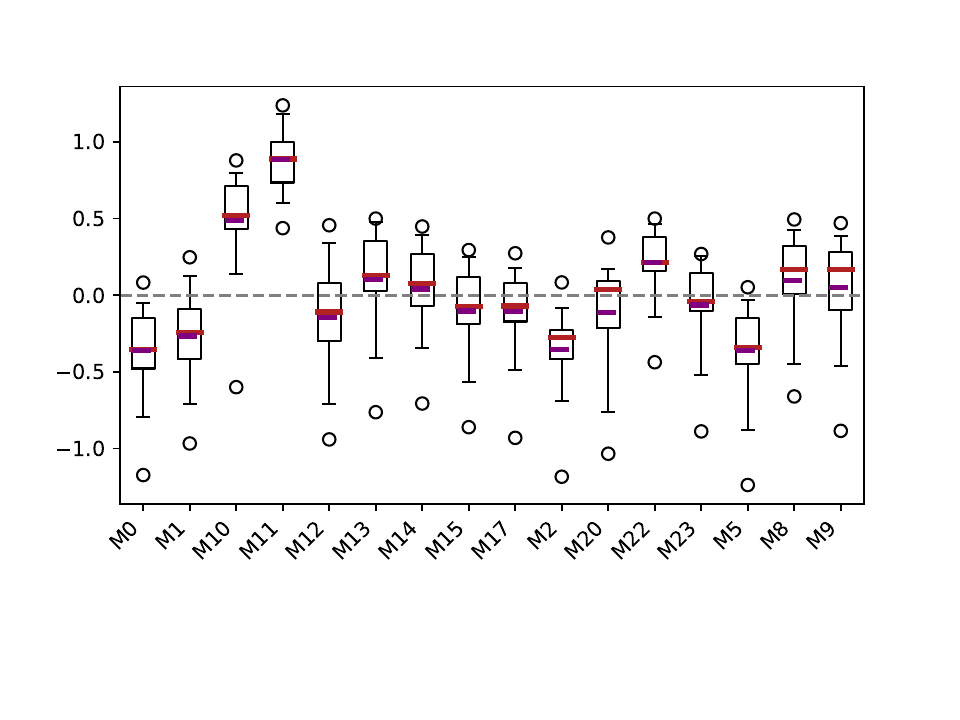}
     \caption{ROUGE}
     \label{fig:ex_rouge_summ_box}
 \end{subfigure}
\caption{ Visualization of the consequences of unfair metrics by comparing the human ranking to the ROUGE ranking. The ranking of a set of systems is depicted as a Directed Acyclic Graph, where an edge from system A to system B states that system A "wins" against system B. Here, a win is determined by a sign test~\cite{sign_tests} at a 95\% confidence threshold.}
\label{fig:ex_dag_summ_rouge}
\end{figure*}

Figure~\ref{fig:ex_dag_summ_rouge} shows the analysis of the ROUGE score. Overall, the ROUGE score has a slightly higher sample-level accuracy (0.07) while having a higher average Favi-Score (0.339), and a much lower system level sign accuracy (0.633). We also note that the favoritism is more pronounced in the ROUGE score, again with \emph{M11} being strongly favored. While \emph{M0} and \emph{M1} are both disfavored, which also is apparent in the final ranking. 

\subsection{Machine Translation}

\begin{figure*}[t!]
\small
\centering
 \begin{subfigure}[c]{0.32\textwidth}
 \centering
     \includegraphics[width=1\textwidth]{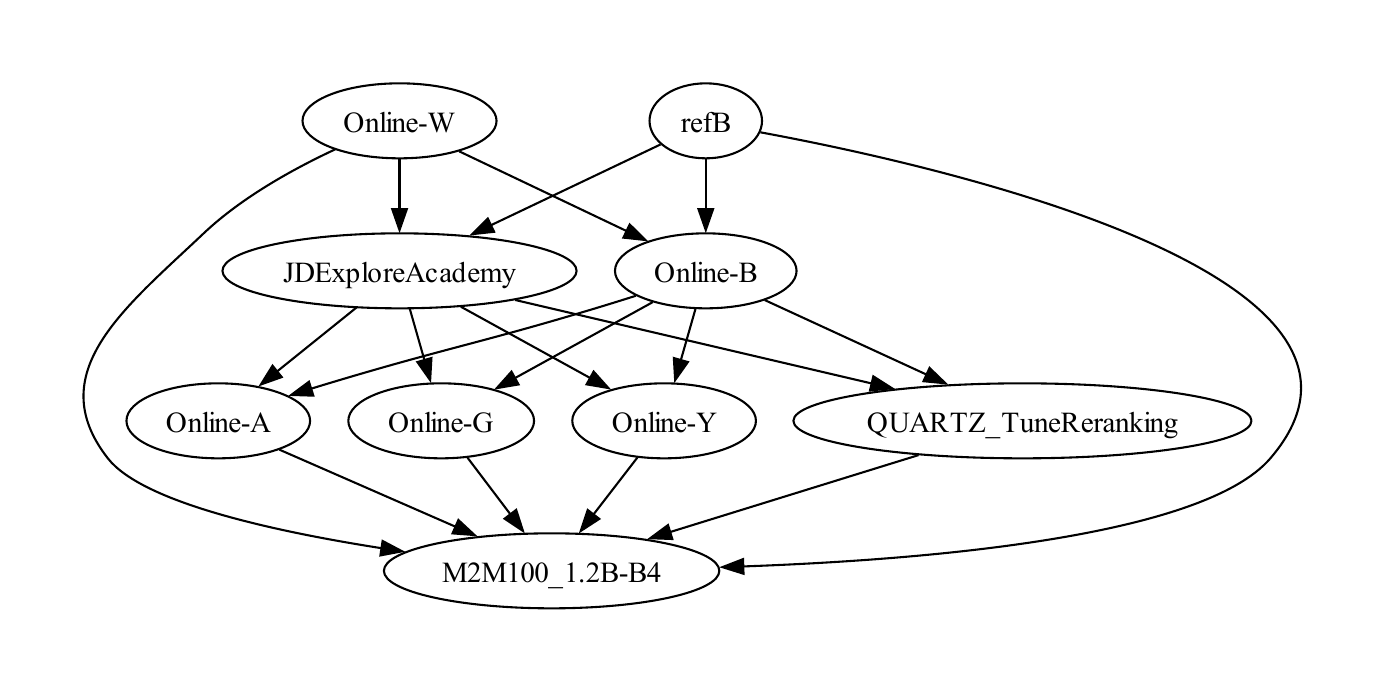}
     \caption{Human}
     \label{fig:ex_chat_mt_hum}
 \end{subfigure}
\begin{subfigure}[c]{0.32\textwidth}
 \centering
     \includegraphics[width=1\textwidth]{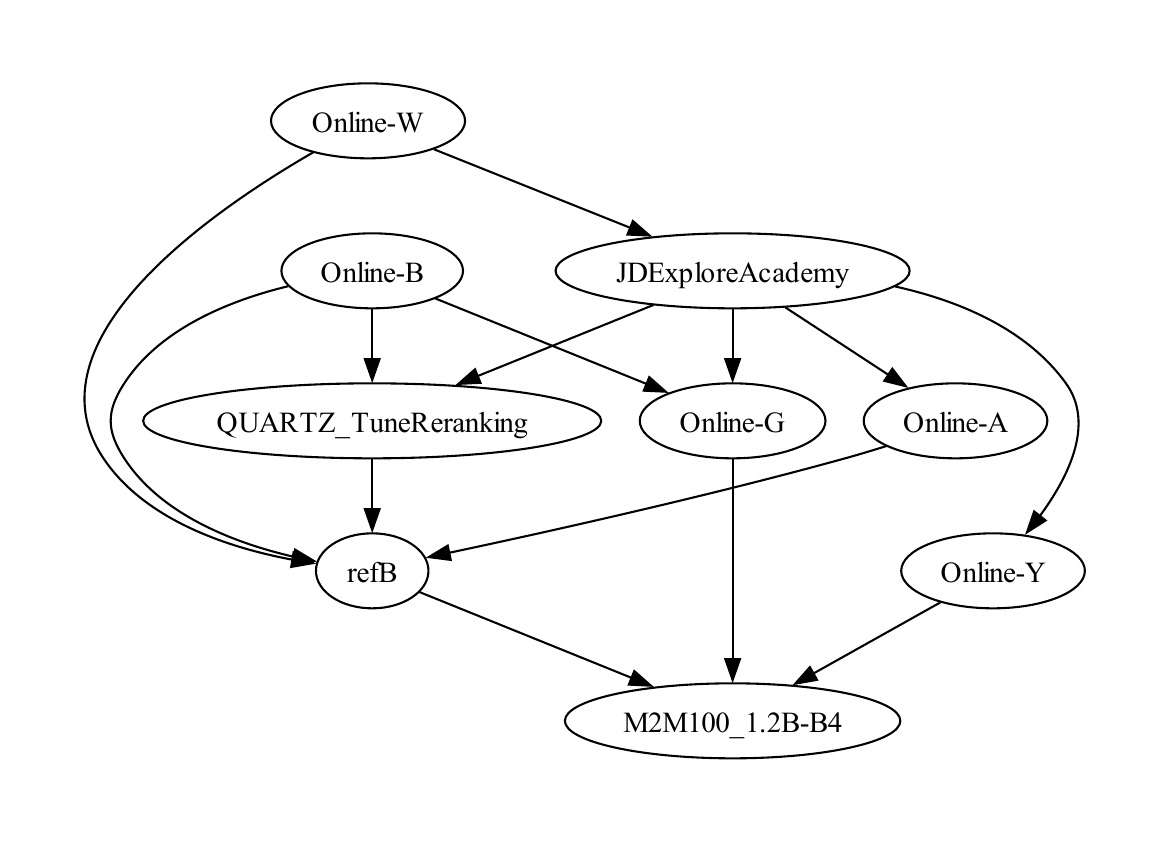}
     \caption{ChatGPT}
     \label{fig:ex_chat_mt_chat}
 \end{subfigure}
 \begin{subfigure}[c]{0.32\textwidth}
 \centering
     \includegraphics[width=1\textwidth]{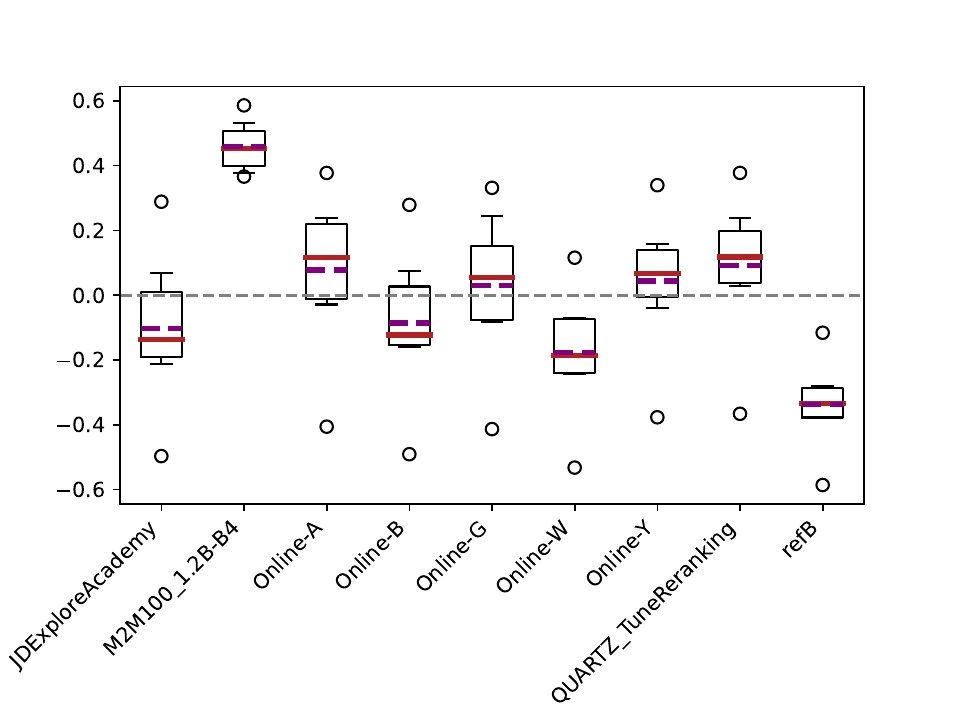}
     \caption{ChatGPT}
     \label{fig:ex_chat_mt_box}
 \end{subfigure}
\caption{ Visualization of the consequences of unfair metrics by comparing the human ranking to the ChatGPT ranking. The ranking of a set of systems is depicted as a Directed Acyclic Graph, where an edge from system A to system B states that system A "wins" against system B. Here, a win is determined by a sign test~\cite{sign_tests} at a 95\% confidence threshold.}
\label{fig:ex_chat_mt}
\end{figure*}

In Figure~\ref{fig:ex_chat_mt} the analysis of ChatGPT as metric is shown. ChatGPT has a high system-level accuracy (0.852), and a relatively low Favi-Score (0.229), as well as a low sample level accuracy (0.348). Among the selected WMT-22 metrics, it has one of the lower system-level accuracy scores, the highest sample-level accuracy score, and the highest Favi-Score. Thus, highlighting the influence of favoritism on the final ranking. Similar to other metrics, it disfavors the human reference (\emph{ref-B}).
It favors \emph{M2M100}, however without chaining the ranking. These are cases where most mistakes are in favor of \emph{M2M100}, however, the outcome margin is too large for the errors to have an impact on the ranking.

\begin{figure*}[t!]
\small
\centering
 \begin{subfigure}[c]{0.32\textwidth}
 \centering
     \includegraphics[width=1\textwidth]{figures/examples/mt/human-human_only-naive.gv.pdf}
     \caption{Human}
     \label{fig:ex_BLEU_mt_hum}
 \end{subfigure}
\begin{subfigure}[c]{0.32\textwidth}
 \centering
     \includegraphics[width=1\textwidth]{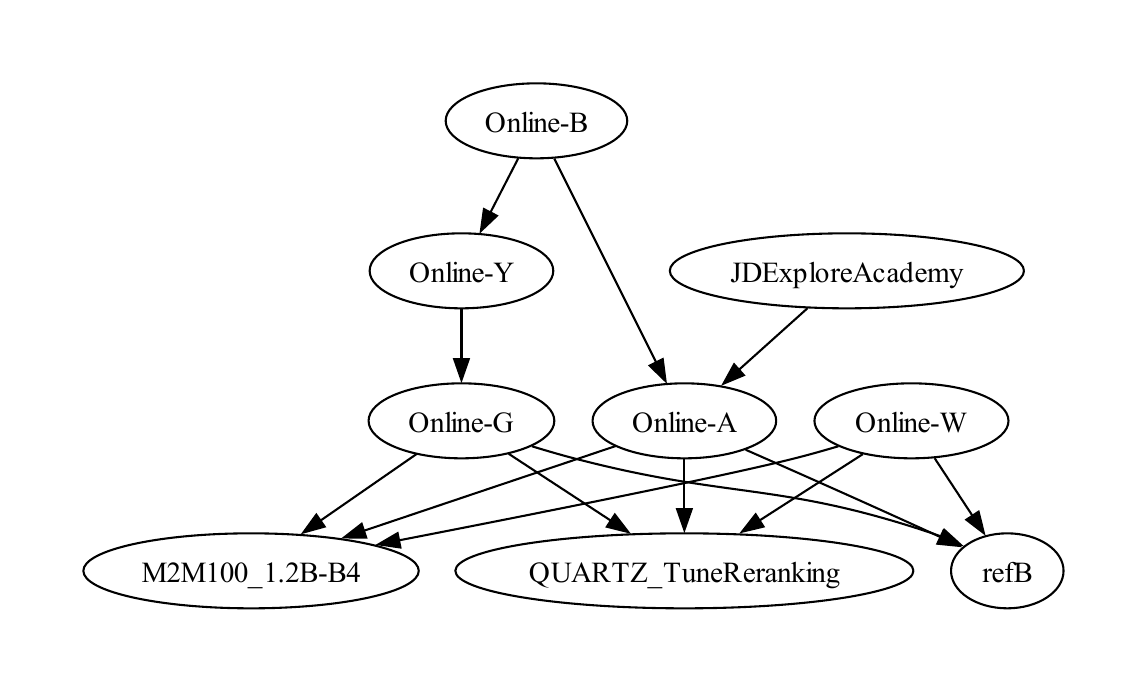}
     \caption{BLEU}
     \label{fig:ex_BLEU_mt_chat}
 \end{subfigure}
 \begin{subfigure}[c]{0.32\textwidth}
 \centering
     \includegraphics[width=1\textwidth]{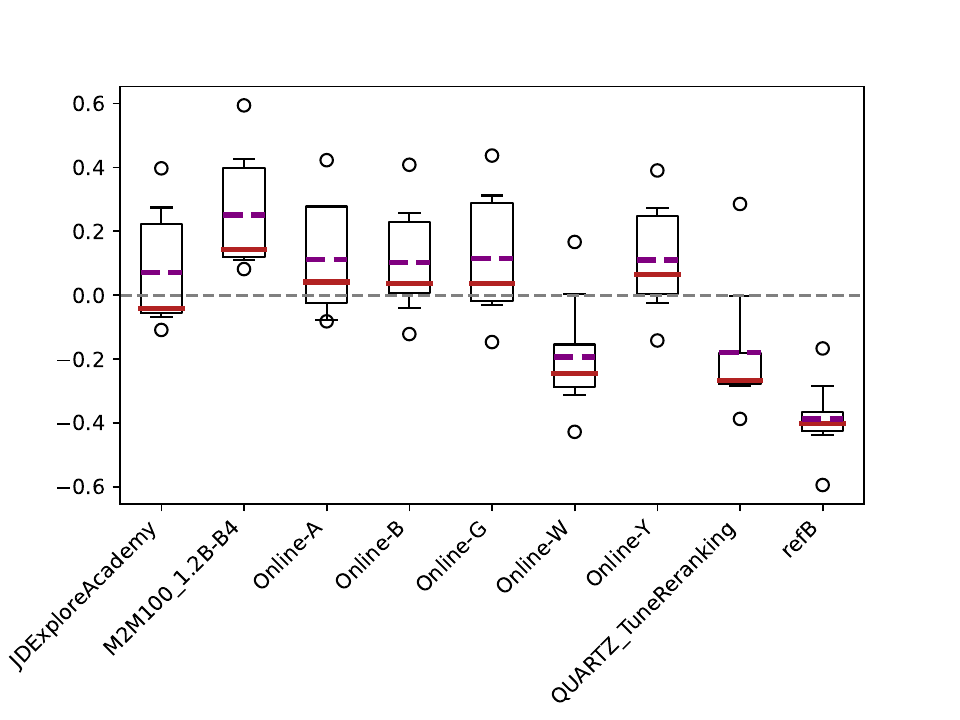}
     \caption{BLEU}
     \label{fig:ex_BLEU_mt_box}
 \end{subfigure}
\caption{ Visualization of the consequences of unfair metrics by comparing the human ranking to the BLEU ranking. The ranking of a set of systems is depicted as a Directed Acyclic Graph, where an edge from system A to system B states that system A "wins" against system B. Here, a win is determined by a sign test~\cite{sign_tests} at a 95\% confidence threshold.}
\label{fig:ex_BLEU_mt}
\end{figure*}

Figure~\ref{fig:ex_BLEU_mt} shows the analysis according to the BLEU score. The BLEU score has the lowest system level sign accuracy (0.7), and one of the higher Favi-Scores (0.21), and sample-level sign accuracy comparable to the other metrics (0.31). Again, we observe that the human reference is rated very poorly, with a high Favi-Score against it. Also \emph{Online-W} has a negative Favi-Score, which also reflects its low position in the ranking according to BLEU. 

\end{document}

%% file: tables/data.tex
\begin{tabular}{ l | c | c | c | c }
&Chatbot&SummEval&WMT21 & WMT22 \\\hline
Data&BST&CNN/DM&News EN->DE&EN->DE \\
Metrics & 5 & 7 & 4 & 5\\
TG Systems & 6 & 16 & 11 & 9\\
|$\mathcal{E}$|& 50 & 100 & 500 & 1315\\
\end{tabular}

%% file: tables/app_proof_enumeration.tex
\begin{center}
\begin{tabular}{|c|c | c| c |c|c || c| c |} \hline
\centering
   $r^H_+$  & $r^H_=$ & $r^H_-$ & $r^A_+$ & $r^A_=$ & $r^A_-$ & LHS & RHS\\ \hline 
   0 & 0 & 1 & 0 & 0 & 1 & 0 & 0 \\ 
   0 & 0 & 1 & 0 & 1 & 0 & 1 & 1 \\ 
   0 & 0 & 1 & 1 & 0 & 0 & 2 & 2 \\
   0 & 1 & 0 & 0 & 0 & 1 & -1 & -1 \\ 
   0 & 1 & 0 & 0 & 1 & 0 & 0 & 0 \\ 
   0 & 1 & 0 & 1 & 0 & 0 & 1 & 1 \\ 
   1 & 0 & 0 & 0 & 0 & 1 & -2 & -2 \\ 
   1 & 0 & 0 & 0 & 1 & 0 & -1 & -1 \\ 
   1 & 0 & 0 & 1 & 0 & 0 & 0 & 0 \\ \hline
\end{tabular}
\end{center}